\documentclass[10pt,twocolumn,letterpaper]{article}

\usepackage{iccv}

\usepackage[margin=6pt,font=small,labelfont=bf,labelsep=endash,tableposition=top]{caption}

\usepackage{graphicx}
\usepackage{amsmath}
\usepackage{amssymb, nicefrac}
\usepackage{float}
\usepackage{subfigure}
\usepackage{graphicx}
\usepackage{comment}

\usepackage{color}
\usepackage{xspace}
\usepackage{textcomp}
\usepackage{cite}
\usepackage{mathtools}
\usepackage{bbm}
\usepackage{makecell, verbatim, sidecap}

\usepackage{multirow}
\usepackage{url,xspace}
\usepackage{xfrac}

\usepackage{pifont}%
\def\eg{{\it e.g.}\xspace}
\def\ie{{\it i.e.}\xspace}

\def\R{{\mathbb R}}

\usepackage[pagebackref=true,breaklinks=true,letterpaper=true,colorlinks,bookmarks=false]{hyperref}

\iccvfinalcopy %

\begin{document}

\title{TFPose: Direct Human Pose Estimation with Transformers}

\author{
Weian Mao$ ^{\dag*}$,  ~~~
Yongtao Ge$ ^{\dag*}$, ~~~
Chunhua Shen$ ^\dag$\thanks{First two authors contributed equally. CS is the corresponding author.}, ~~~
Zhi Tian$ ^\dag$, ~~~
Xinlong Wang$ ^\dag$, ~~~
Zhibin Wang$ ^\ddagger$
\\[0.2cm] 
$ ^\dag$ The University of Adelaide
~~~~ ~~~~ ~~~~
$ ^\ddagger$ Alibaba Group
}

\maketitle
\ificcvfinal\thispagestyle{empty}\fi

\begin{abstract}
We propose a human pose estimation framework that solves the task in the regression-based fashion. Unlike previous regression-based methods, which often %
fall behind those state-of-the-art methods, 
we formulate the pose estimation task into a 
sequence 
prediction problem that can effectively be solved by transformers. Our framework is simple and direct,
bypassing %
the drawbacks of the heatmap-based pose estimation. Moreover, with the attention mechanism in transformers, our proposed framework is able to adaptively attend to the features most relevant to the target keypoints, which
largely 
overcomes the feature misalignment issue of previous regression-based methods and considerably improves the performance.
Importantly, 
our framework can inherently take advantages of the structured relationship between keypoints.
Experiments on the MS-COCO and MPII datasets demonstrate that our method can significantly improve the state-of-the-art of regression-based pose estimation and perform comparably with the best heatmap-based pose estimation methods.

 {   
    \def\UrlFont{\sf}
    \def\UrlFont{\rm\small\ttfamily}
Code is available at: \url{https://git.io/AdelaiDet} 
}

\end{abstract}

\section{Introduction}

Human pose estimation requires the computer to obtain the human keypoints of interest in an input image and plays an important role in many computer vision tasks such as human behavior understanding.

\begin{figure}[t]
\centering 
\subfigure[Heatmap-based method]{
	\includegraphics[width=0.9\columnwidth]{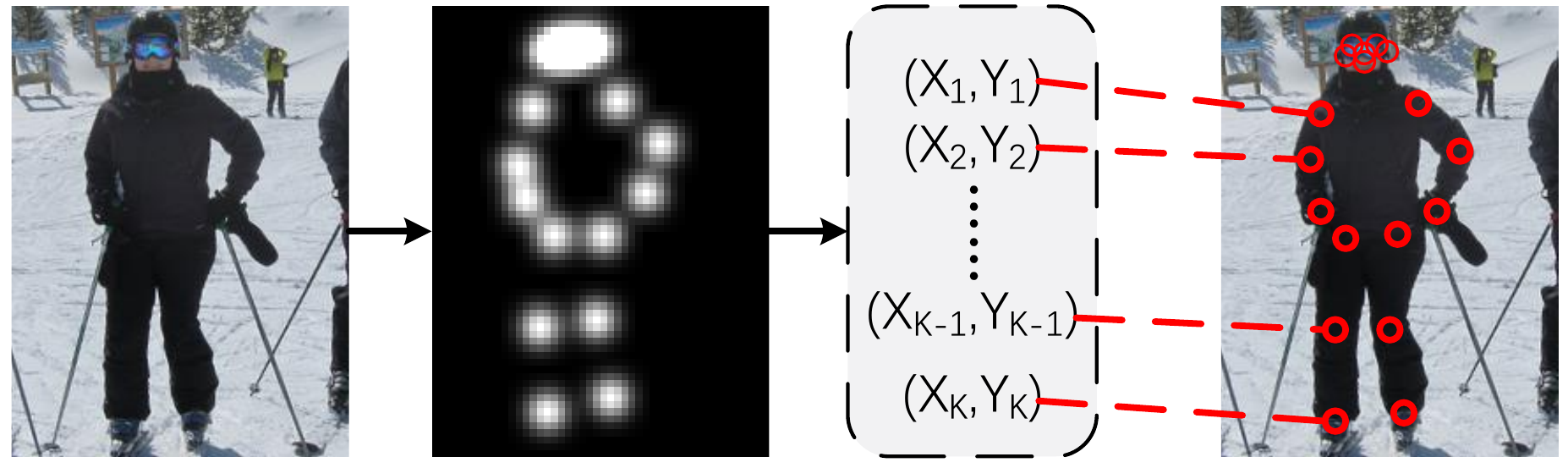}
	}
\subfigure[Regression-based method]{
	\includegraphics[width=0.9\columnwidth]{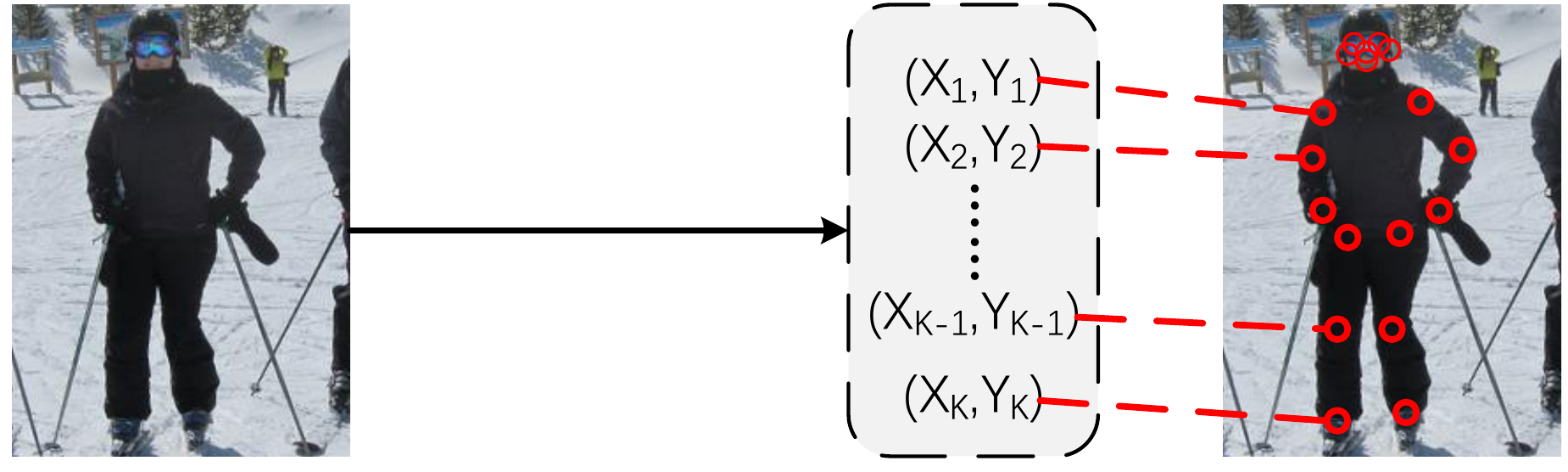}
	}
\caption{Comparison of mainstream pose estimation pipelines. (a) Heatmap-based methods. (b) Regression-based methods. 
}
\label{fig:difference}
\end{figure}

Existing mainstream methods solving the task can be generally categorized into heatmap-based (Figure~\ref{fig:difference} top) and regression-based methods (Figure.~\ref{fig:difference} bottom).
Heatmap-based methods often first predict a heatmap or a classification score map with fully convolutional networks (FCNs), and then the body joints are located by the peak's locations in the heatmap or the score map. Most pose estimation methods are heatmap-based because it has relatively higher accuracy. However, the heatmap-based methods may suffer the following issues. 1) A post-processing (\eg, the ``taking-maximum'' operation) is needed. The post-processing might not be differentiable, making the framework not end-to-end trainable. 2) The resolution of heatmaps predicted by the FCNs is usually lower than the resolution of the input image. The reduced resolution results in a quantization error and limits the precision of the keypoint's localization. This quantization error might be solved by shifting the output coordinates according to the value of the pixels near the peak, but it makes the framework much more complicated and introduces more hyper-parameters. %
3) The ground truth heatmaps need to be manually designed and heuristically tuned, which might cause many noises and ambiguities contained in the ground-truth maps, as show in \cite{luo2020rethinking, sun2017compositional, huang2020devil}. 

In contrast, the regression-based methods usually directly map the input image to the coordinates of body joints with a FC (fully-connected) prediction layer, eliminating the need for heatmaps. The pipeline of regression-based methods is much more straightforward than heatmap-based methods as in principle pose estimation is a kind of regression tasks such as object detection. Moreover, the regssion-based method can bypass the aforementioned drawbacks of heatmap-based methods, thus being more promising. %

However, there are only a few research works focusing on regression-based methods because regression-based methods often have inferior performance to heatmap-based methods. The reasons may be four-fold. First, in order to reduce the network parameters in the FC layer, in the DeepPose~\cite{toshev2014DeepPose}, a global average pooling is applied to reduce the feature map resolution before the FC layer. This global average pooling destroys the spatial structure of the convolutional feature maps, and significantly deteriorates the performance. Next, as shown in DirectPose~\cite{tian2019directpose} and SPM~\cite{nie2019single}, in regression-based methods, the convolutional features and predictions are misaligned, which results in low localization precsion of the keypoints. Moreover, regression-based methods only regress the coordinates of body joints and does not take account of the structured dependency between these keypoints\cite{sun2017compositional}. %

\begin{figure*}[h]
\centering 
\includegraphics[width=0.9585\textwidth]{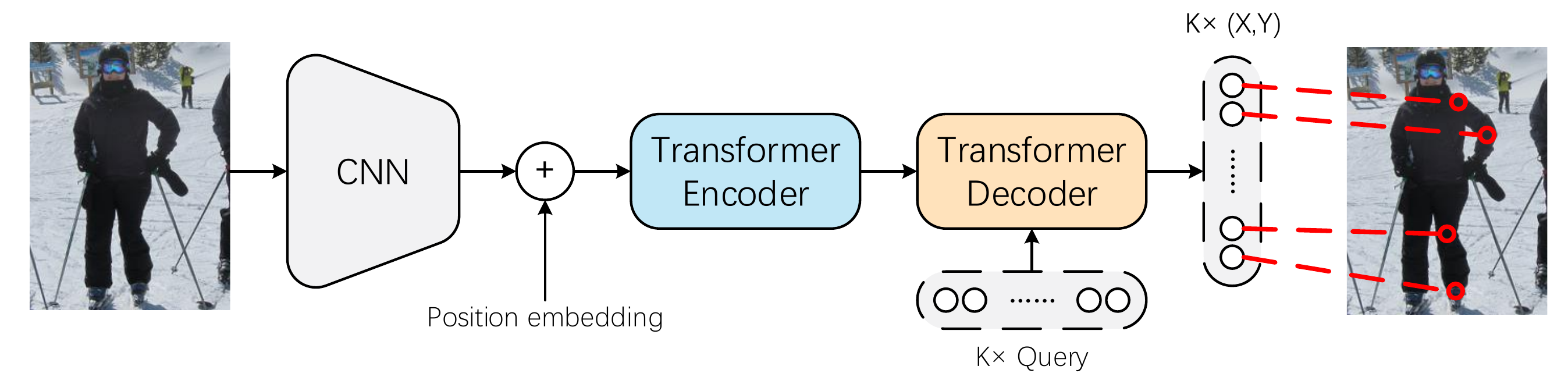}
\caption{\textbf{Overall pipeline of TFPose.} The model directly predicts a sequence of keypoint coordinates in parallel by combining a common CNN with a transformer architecture. A transformer decoder takes as input a fix number of keypoint queries and encoder output. Then, we pass the output embedding of the decoder to a multi-layer feed forward network that predicts final keypoint coordinates.}
\label{fig:framework}
\end{figure*}

Recently, we have witnessed the rise of vision transformers~\cite{zhu2020deformable, dosovitskiy2020image, carion2020end}. The transformers are originally designed for the sequence-to-sequence tasks, which inspires us to formulate the single person pose estimation to the problem of predicting K-length sequential coordinates, where K is the number of body joints for one person. This leads to a simple and novel regression-based pose estimation framework, termed TFPose (\ie, Transformer-based Pose Estimation). As shown in Figure~\ref{fig:framework}, taking as inputs the feature maps of CNNs, the transformer sequentially predict $K$ coordinates. %
TFPose can bypass the aforementioned difficulties of regression-based methods. First, it does not need the global average pooling as in DeepPose~\cite{toshev2014DeepPose}. Second, due to the multi-head attention mechanism, our method can avoid the feature misalignment between the convolutional features and predictions. Third, since we predict the keypoints in the sequential way, the transformer can naturally capture the structured dependency between the keypoints, resulting in improved performance. %

We summarize the main contributions as follows.
\begin{itemize}
    \itemsep 0cm 

\item  TFPose is the first transformer-based pose estimation framework. Our proposed framework adapts to the simple and straightforward regression-based methods, which is end-to-end trainable and can overcome many drawbacks of the heatmap-based methods.

\item  Moreover, our TFPose can naturally learn to 
exploit 
the structured dependency between the keypoints without heuristic designs,  \textit{e.g.}, in \cite{sun2017compositional}. This results in improved performance and better interpretability.

\item 
TFPose achieves greatly advance the state-of-the-art of regression-based methods, making the regression-based methods comparable to the state-of-the-art heatmap-based ones. For example, we improve the previously best regression-based method Sun et al.~\cite{sun2018integral} by 4.4\% AP on the COCO keypoint detection task, and Aiden et al.~\cite{nibali2018numerical} by 0.9\% PCK on the MPII benchmark.
\end{itemize}

\section{Related Work}
\noindent\textbf{Transformers in computer vision.}
After being proposed in~\cite{vaswani2017attention}, Transformers have achieved significant progress in NLP (Natural Language Processing)~\cite{devlin2018bert, brown2020language}.
Recently, Transformers have also attracted much attention in computer vision community.
For basic image classification task, ViT\cite{dosovitskiy2020image} apply a pure Transformer to sequential image patches. Expect for image classification, vision Transformer is also widely applied to object detection\cite{carion2020end,zhu2020deformable}, segmentation\cite{wang2020max, wang2020end}, pose estimation\cite{huang2020hand,huang2020hot, lin2020end}, low-level
vision task \cite{chen2020pre%
}. More details, we refer to \cite{han2020survey}. Specially, DETR\cite{carion2020end} and Deformable DETR\cite{zhu2020deformable} formulate the object detection task to predict a box set so that object detection model can be trained end-to-end; the Transformer applications in both 3D Hand Pose Estimation\cite{huang2020hand,huang2020hot} and 3D human pose estimation\cite{huang2020hand,huang2020hot} show that Transformer is suitable for modeling human pose.

\noindent\textbf{Heatmap-based 2D pose estimation.}
Heatmap-based 2D pose estimation methods\cite{chen2018cascaded, xiao2018simple, sun2019deep, cai2020learning, li2019rethinking, cao2019openpose, cheng2020higherhrnet, he2017mask, newell2016stacked} perform the state-of-the-art accuracy in 2D human pose estimation. Recently, most work, including both top-down and bottom up, are heatmap-based methods. \cite{newell2016stacked} firstly propose hourglass-style framework and hourglass-style framework also be widely applied, such as, \cite{cai2020learning, li2019rethinking}. \cite{sun2019deep} propose a novel network architecture for heatmap-based 2D pose estimation and achieve a excellent performance. \cite{cheng2020higherhrnet} propose a new bottom-up method achieve impressive performance in CrowdPose dataset\cite{li2019crowdpose} and improved by \cite{luo2020rethinking}. \cite{cai2020learning} propose a efficient network achieving the the-state-of-art performance in COCO keypoint detection dataset\cite{lin2014microsoft}. However, \cite{sun2018integral, tian2019directpose} argue that heatmap-based methods cannot be trained end-to-end, due to the "taking-maximum'' operation. Recently, the noise and ambiguity in the ground truth heatmap are found by \cite{luo2020rethinking, sun2017compositional}. \cite{huang2020devil} finds the heatmap data processing applied by most previous work is biased and proposed an new unbiased data processing method.

\noindent\textbf{Regression-based 2D pose estimation.}
2D human pose estiamtion is naturally a regression problem\cite{sun2018integral}. However, regression based methods are not accurate as well as heatmap-based methods, thus there are just a few works\cite{toshev2014DeepPose, sun2018integral, sun2017compositional,  carreira2016human, tian2019directpose, nie2019single} for it. Apart from that, although some methods, such as G-RMI \cite{papandreou2017towards}, apply regression method to reduce the quantization errors casued by heatmap, they are essentially heatmap-based methods. There are some work point out the reason of the bad performance of regression-based method. Directpose\cite{tian2019directpose} points out the feature mis-alignment issue and propose a mechanism to align the feature and the predictions; \cite{sun2017compositional} indicates regression-based method cannot learn the structure-aware information well and proposal a hand-design model for pose estimation to force regression-based method learn the structure-aware information better; Sun et al.\cite{sun2018integral} propose integral regression, which shares the merits of both heatmap representation and regression approaches, to avoid non-differentiable postprocessing and quantization error issues.

\section{Our Approach}

\subsection{TFPose Architecture}\label{sec:TFPose Architecture}
This work focus on the single pose estimation task. Following previous works, we first apply a person detector to obtain the bounding boxes of persons. Then, according to the detected boxes, each person is cropped from the input image. We denote the cropped image by $\textbf{I} \in \R^{h \times w \times 3}$, where $h$, $w$ is the height and the width of the image, respectively. With the cropped image with a single person, the previous heatmap-based methods apply a convolutional neural network $\mathcal{F}$ to the patch to predict keypoint heatmaps $\textbf{H} \in \R^{h \times w \times k}$($H_k$ for $k^{th}$ joint) of this person, where $k$ is the number of the predicted keypoint. Formally, we have
\begin{equation}
    \textbf{H} = \mathcal{F}(\textbf{I}).
\end{equation}
Each pixel of $\textbf{H}$ represents the probability that the body joints locate at this pixel. To obtain the joints' coordinates $\textbf{J} \in \R^{2 \times k}$($\textbf{J}_k$ for $k^{th}$joint), those methods usually use the ``taking-maximum" operation to obtain the locations with peak activations. Formally, let $\textbf{p}$ be the spatial locations on $\textbf{H}$, and it can be formulated as
\begin{equation}\label{equation:max op}
    \textbf{J}_k = \mathop{\arg\max}_{\textbf{p}}(\textbf{H}_k(\textbf{p})).
\end{equation}
Note that in the heatmap-based methods, the localization precision of $\textbf{p}$ is up to the resolution of \textbf{H}, which is often much lower than the resolution of the input and thus causes the quantization errors. Moreover, the $\mathop{\arg\max}$ operation here is not differential, making the pipeline not end-to-end trainable. In TFPose, we instead treat $\textbf{J}$ as a K-length sequence and directly map the input $\textbf{I}$ to the body joints' coordinates $\textbf{J}$. Formally,
\begin{equation}\label{equation:max}
    \textbf{J} = \mathcal{F}(\textbf{I}),
\end{equation}
where $\mathcal{F}$ is composed of three main components: a standard CNN backbone to extract multi-level feature representations, a feature encoder to capture and fuse multi-level features 
and a coarse-to-fine decoder to generate the a sequence of keypoint coordinates. It is illustrated in Figure~\ref{fig:framework}. Note that our TFPose is fully differentiable and the localization precision is not limited by the resolution of the feature maps.

\subsection{Transformer Encoder} \label{section: encoder}

\begin{figure}[b]
\centering 
\includegraphics[width=0.40\textwidth]{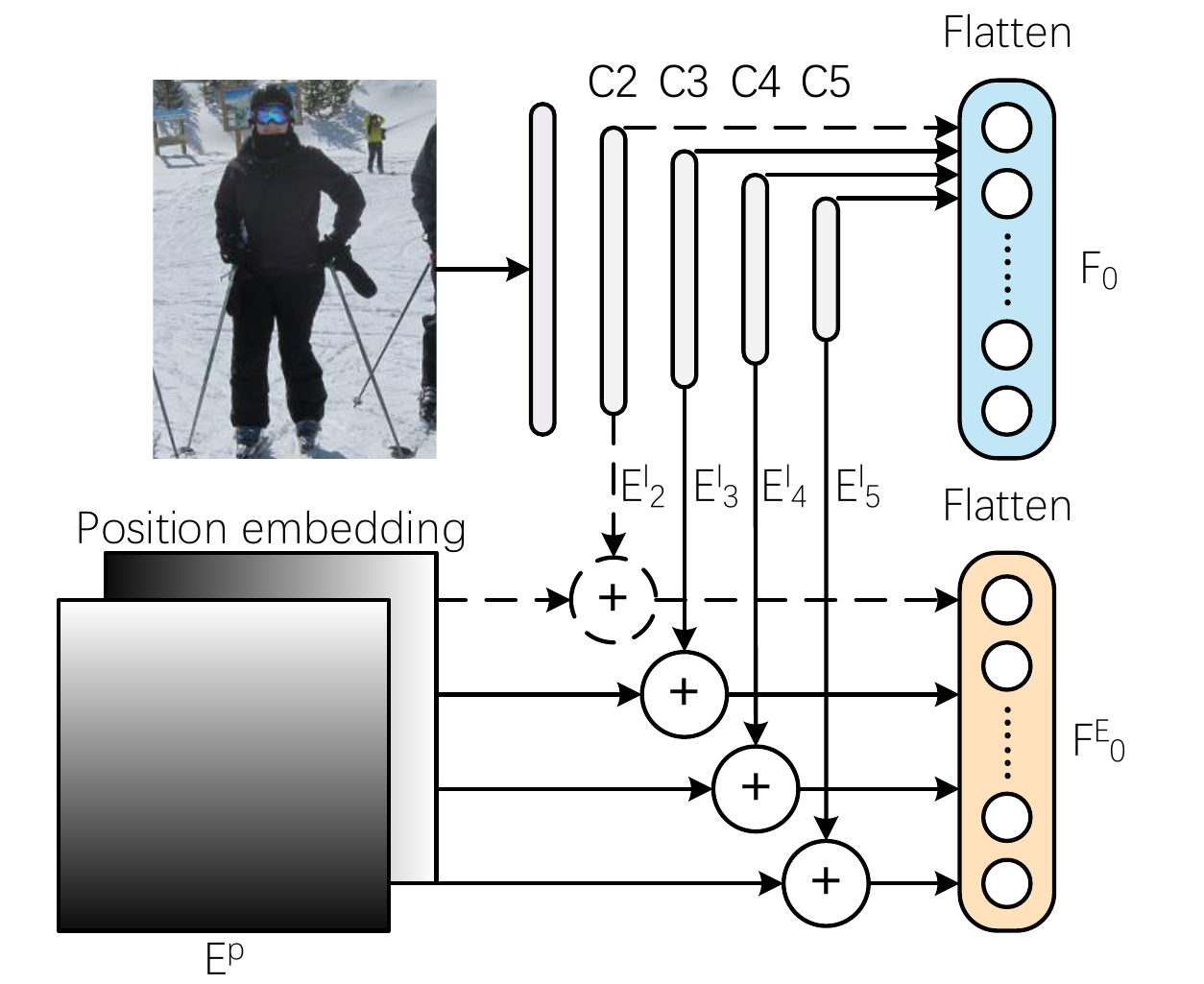}
\caption{\textbf{Positional encoding.} This figure illustrates the positional embeddings to the input $F_0$ of the transformer. $E^l_i$ represents the level embeddings depicting which level a feature vector comes from. $E^P$ represents the pixel embedding depicting the spatial location of a feature vector on the feature maps. We use $F^E_0$ to denote $F_0$ with position embedding. Following ~\cite{zhu2020deformable}, both $F_0$ and $F^E_0$ are the inputs of the transformer.}
\label{fig:pos_emb}
\end{figure}

As shown in Figure~\ref{fig:pos_emb}, the backbone extracts multi-level features of the input image. The multi-level feature maps are denoted by $C_2$, $C_3$, $C_4$ and $C_5$, respectively, whose strides are 4, 8, 16 and 32, respectively. We separately apply a $1\times1$ convolution to these feature maps so that they have the same number of the output channels. These feature maps are flatten and concatenated together, which results in the input $F_0 \in \R^{n \times c}$ to the first encoder in the transformer , where $n$ is the number of the pixel in the $F_0$. Here, we use $F_i$ denotes the output to the $i$-th encoder in the transformer. Following \cite{zhu2020deformable, vaswani2017attention}, $F_0$ is added with the positional embeddings and we denote $F_0$ with the positional embeddings by $F_0^E$. The details of the positional embeddings will be discussed in Section \ref{section: encoder}. Afterwards, both $F_0$ and $F_0^E$ are sent to the transformer to compute the memory $M\in \R^{n \times c}$. With the memory $M$, a query matrix $Q \in \R^{K \times c}$ will be used in the transformer decoder to obtain the K body joints' coordinates $\textbf{J} \in \R^{K \times 2}$. %

We follow Deformable DETR~\cite{zhu2020deformable} to design the encoder in our transformer. As mentioned before, before $F_0$ is taken as inputs, each feature vector of $F_0$ is added with the positional embeddings. Following Deformable DETR, we use both level embedding $E^L_l \in \R^{1 \times c}$ and pixel position embeddings $E^P \in \R^{n \times c}$. The former encodes the level where the feature vector comes from, and the latter is the feature vector's spatial location on the feature maps. As shown in Figure~\ref{fig:pos_emb}, all the feature vectors from level $l$ are added with $E^L_l$ and then the feature vectors are added with their pixel position embeddings $E^p$, where $E^p$ is the 2-D cosine positional embeddings corresponding to the 2-D location of the feature vector on the feature maps.
In TFPose, we use $N_E = 6$ encoder layers. For $e^{th}$ encoder layer, as shown in Figure~\ref{fig:transformer}, the previous encoder layer's outputs will be taken as the input of this layer. Following Deformable DETR, we also compute the pixel-to-pixel attention between the output vectors of each encoder layer (denoted by `p2p attention'). %
After $N_E$ transformer encoder layers are applied, we can obtain the memory $M$.

\subsection{Transformer Decoder}\label{sec: Transformer Decoder} 
\begin{figure}[b]
\centering 
\includegraphics[width=0.40\textwidth]{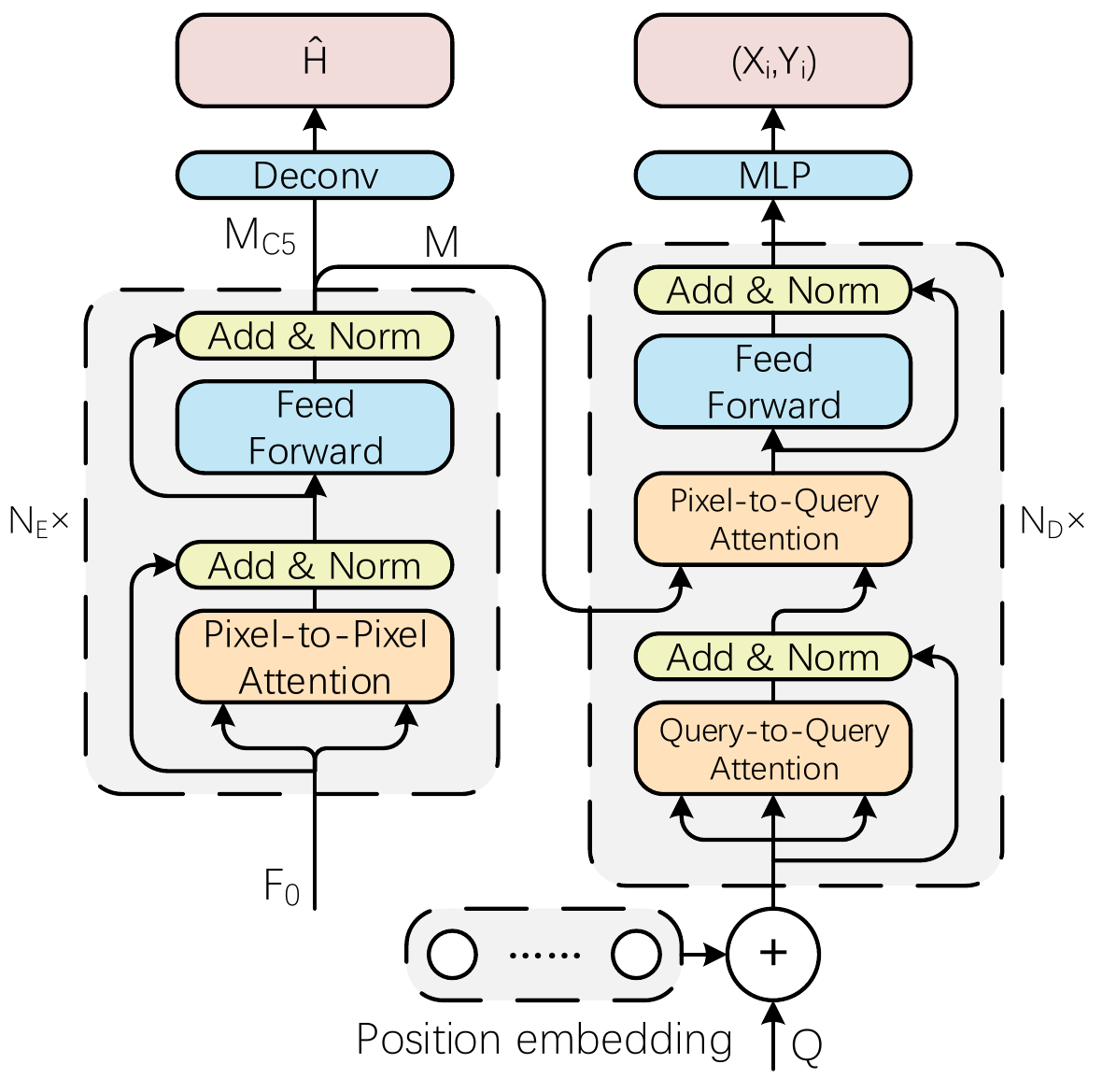}
\caption{\textbf{Transformer architecture.} During training, deconvolution modules are used to upsamle transformer encoder output ($M_{C_5}$) for for auxiliary loss. During testing, only output Transformer decoder. `Norm' represent normalization; ($X_i$, $Y_i$) represent the coordinate for $i^{th}$ keypoint.}
\label{fig:transformer}
\vspace{-0.5em}
\end{figure}
In the decoder, we aim to decode the desired keypoint coordinates from the memory $M$. As mentioned before, we use a query matrix $Q \in \R^{K \times c}$ to achieve this. $Q$ is essentially an extra learnable matrix, which is jointly updated with the model parameters during training and each row of which corresponds to a keypoint. In TFPose, we have $N_D$ transformer decoder layers. As shown in Figure \ref{fig:transformer}, each decoder layer takes as input the memory $M$ and the outputs of the previous decoder layer $Q_{d-1} \in \R^{K \times c}$. The first layer takes as inputs $M$ and the matrix $Q$. Similarly, $Q_{d-1}$ is added with the positional embeddings. The result is denoted by $Q_{d-1}^E$. The $Q_{d-1}$ and $Q_{d-1}^E$ will be sent to the query-to-query attention module (denoted as `q2q attention'), which aims to model the dependency  between human body joints. The q2q attention module use $Q_{d-1}$, $Q_{d-1}^E$ and $Q_{d-1}^E$ as values, queries and keys, respectively. 
Later, the output of the q2q attention module and $M$ used to compute the pixel-to-query attention (denoted as `p2q attention') with the value being the former and query being the latter.
Then, an MLP will be applied to the output of p2q attention the output of the decoder $Q_{d}$. The keypoint coordinates are predicted by applying an MLP with output channels being $2$ to each row of $Q_{d}$.

Instead of simply predicting the keypoint coordinates in the final decoder layer, inspired by \cite{carreira2016human, hu2018facial, zhu2020deformable}, we require all the decoder layers to predict the keypoint coordinates. Specifically, we let the first decoder layer directly predict the target coordinates. Then, every other decoder layer refines the predictions of its previous decoder layer by predicting refinements $\Delta \hat{y}_{d} \in \R^{K \times 2}$. In that way, the keypoint coordinates can be progressively refined. Formally, let $\hat{y}_{d-1}$ be the keypoint coordinates predicted by the $(d-1)$-th decoder layer, the predictions of the $d$-th decoder layer are
\begin{eqnarray}
\hat{y}_{d}={\sigma(\sigma^{-1}(\hat{y}_{d-1}) + \Delta \hat{y}_{d})},
\label{eq:ddd}
\end{eqnarray}
where $\sigma$ and $\sigma^{-1}$ denote the sigmoid and inverse sigmoid function, respectively. $\hat{y}_0$ is a randomly-initialized matrix and jointly updated with model parameters during training.

\subsection{Training Targets and Loss Functions}
The loss functions of TFPose consist of two parts. The first part is the $\mathcal{L}_1$ regression loss.
Let $y \in \R^{K \times 2}$ be the ground-truth coordinates. The regression loss is formulated as,

\begin{equation}\label{equation:reg loss func}
    L_{reg} = \sum_{d=1}^{N_D} ||{y}-\hat{{y}}_d||,
\end{equation}

where $N_D$ is the number of the decoders, and every decoder layer is supervised with the target keypoint coordinates. The second part is an auxiliary loss $L_{aux}$. Following DirectPose~\cite{tian2019directpose}, we use the auxiliary heatmap learning during training \footnote{The heatmap branch is removed in inference.}, which can result in better performance. In order to use the heatmap leanrning, we gather the feature vectors that were $C_5$ from $M$ and reshape these vectors into the original spatial shape. The result is denoted by $M_{C_5} \in \R^{(h/32) \times (w/32) \times c}$. Similar to simple baseline\cite{xiao2018simple}, we apply $3\times$ deconvolution to $M_{C_5}$ to upsample the feature maps by $8$ and generate the heatmap $\hat{H} \in \R^{(h/4) \times (w/4) \times K}$. Then, we compute the mean square error (MSE) loss between the predicted and ground-truth heatmaps. The ground-truth heatmaps are generated by following \cite{newell2016stacked, Xiao_2018_ECCV}. Formally, the auxiliary loss function is
\begin{equation}
    L_{aux} = ||H - \hat{H}||^2,
\end{equation}

We sum the two loss functions to obtain the final overall loss

\begin{equation}\label{equation:loss func}
    L_{overall} = L_{reg} + \lambda L_{aux},
\end{equation}
where $\lambda$ is a constant and used to balance the two losses.

\section{Experiments}

\subsection{Implementation details.}
\noindent\textbf{Datasets.} 
We conduct a  number of ablation experiments on two mainstream pose estimation datasets.

Our experiments are mainly conducted on COCO2017 Keypoint Detection\cite{wang2018mscoco} benchmark, which contains about $250K$ person instances with 17 keypoints. 
Following common settings, we use the same person detector in Simple Baseline~\cite{Xiao_2018_ECCV} for COCO evaluation. We report results on the \texttt{val} set for ablation studies and compare with other state-of-the-art methods on the \texttt{test-dev} set. The Average Precision (AP) based on Object Keypoint Similarity (OKS) is employed as the evaluation metric. %

Besides COCO dataset, we also report results on MPII dataset~\cite{mpii}.
MPII is a popular benchmark for single person 2D pose estimation, which has $25K$ images. 
In total, there are $29K$ annotated poses for training, and another $7K$ poses for testing. 
The Percentage of Correct Keypoints (PCK) metric is used for evaluation. 

\noindent\textbf{Model settings.} 
Unless specified, ResNet-18\cite{he2016deep} is used as the backbone in ablation study. 
The size of input image is $256\times192$ or $384\times288$.
The weights pre-trained on ImageNet\cite{deng2009imagenet} are used to initialize the ResNet backbone. The rest parts of our network are initialized with random parameters.  For the Transformer, we adopt Defermable Attention Module proposed in~\cite{zhu2020deformable} and the same hyper-parameters are used. 

\noindent\textbf{Training.}  All the models are optimized by AdamW\cite{loshchilov2017decoupled} with a base learning rate of $4\times10^{-3}$.
$\beta_1$ and $\beta_2$ are set to $0.9$ and $0.999$.
Weight decay is set to $10^{-4}$. 
$\lambda$ is set to $50$ by default for balancing the regression loss and auxiliary loss. 
Unless specified, all the experiments use a cosine learning schedule with base learning rate $4\times10^{-3}$.
Learning rate of the Transformers and the linear projections for predicting keypoints offsets is decreased by a factor of $10$. For data augmentation,  random rotation ($[-40^{\circ},40^{\circ}]$), random scale ($[0.5,1.5]$), flipping and half body data augmentation\cite{wang2018mscoco} are applied. For auxiliary loss, we follow Unbiased Data Processing (UDP) \cite{huang2020devil} to generate unbiased ground truth.

\subsection{Ablation Study}

\begin{figure}[t]
\begin{center}
\includegraphics[width=0.45\textwidth]{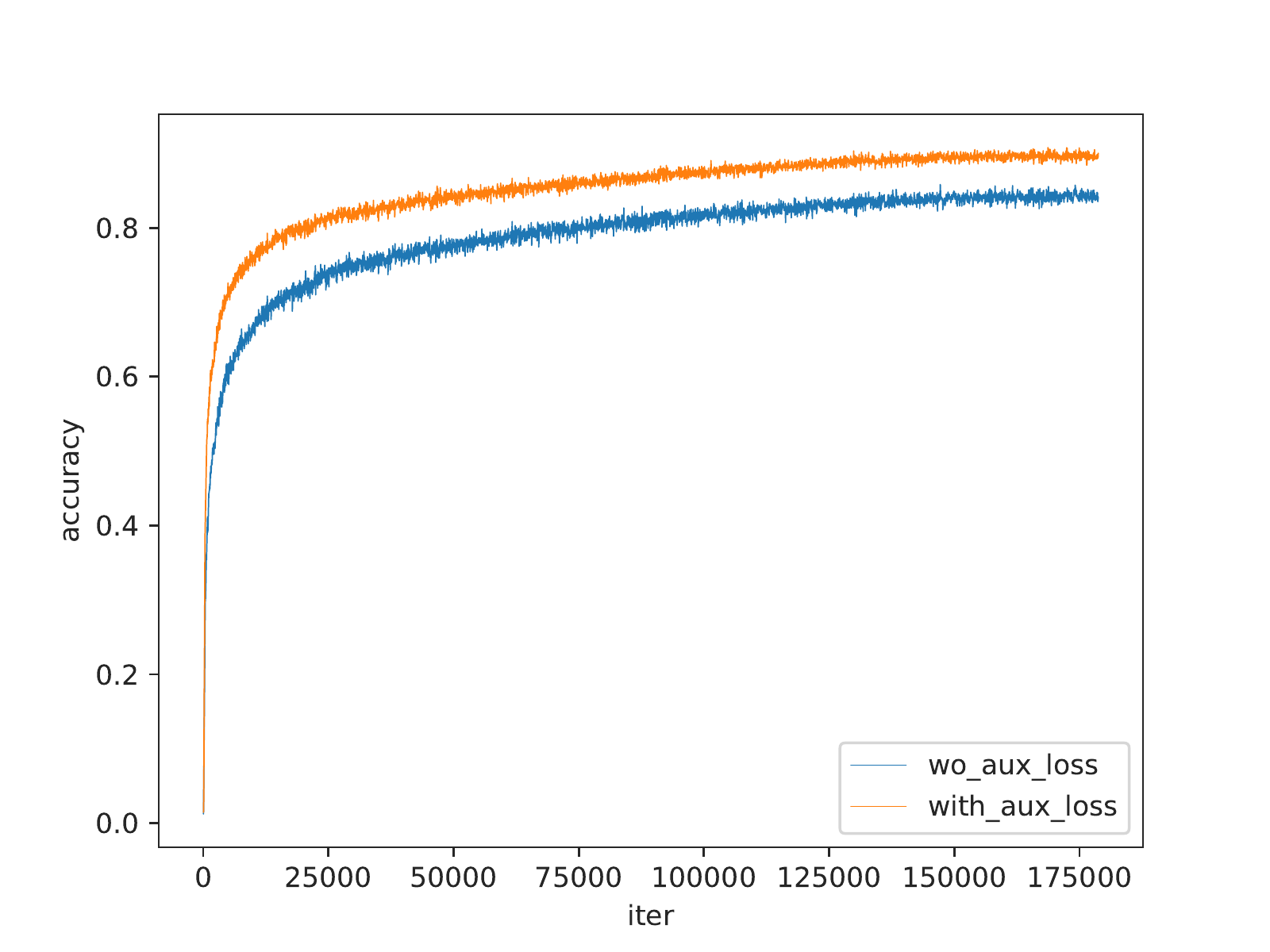}
\end{center}
  \caption{\textbf{Convergence curves of TFPose} superivsed by different kinds of losses on COCO \emph{val2017} set. "{wo aux loss}" indicates only regression loss is employed. "with aux loss" indicates both regression loss and auxiliary loss are employed.}
\label{fig:aux loss}
\end{figure}

\noindent\textbf{Query-to-query attention.}
In the proposed TFPose, query-to-query attention is designed to capture structure-aware information cross all the keypoints. Unlike \cite{sun2017compositional} which uses a hand-design method to explicitly force the model to learn the structure-aware information, query-to-query attention models human body structure implicitly. 
To study the effect of query-to-query attention, we report the results of removing the query-to-query attention in all decoder layers. 
As shown in Table~\ref{table: comparison basline with basline without q2q self attention}, the proposed query-to-query attention improve the performance by 1.3\% AP with only 0.1 GFLOPs more computational cost. 

\begin{table}[t]
    \small
    \centering
	\begin{tabular}{c|c|c|cc|cc}
	q2q & GFLOPs & AP$^{kp}$ & AP$^{kp}_{50}$ & AP$^{kp}_{75}$ & AP$^{kp}_{M}$ & AP$^{kp}_{L}$ \\
	\hline
	 & 3.51 & 63.2 & 85.1 & 69.9 & 60.3 & 70.2 \\
	 \checkmark & 3.61 & \textbf{64.5} & \textbf{85.2} & \textbf{71.2} & \textbf{61.5} & \textbf{71.5} \\
	\end{tabular}
	\vspace{0.5em}
	\caption{The effect of query-to-query attention in decoder layers on COCO \emph{val2017} set. "q2q" indicates whether add query-to-query attention in the decoder. In this experiment, we set the number of transformer encoder layers: $N_E=0$, and transformer decoder layers: $N_D=6$. As shown in the table, decoder with query-to-query attention have better performance.}
	\label{table: comparison basline with basline without q2q self attention}
\end{table}

\noindent\textbf{Configurations of Transformer decoder.}
Here we study the effect of width and depth of the decoder.
Specifically, we conduct experiments by varying the number of channels of the input features and the number of decoder layers in Transformer decoder. 

As shown in Table~\ref{table: comparison Transformer chennal},  
Transformers with 256-channel feature maps is 1.3\% AP higher than 128-channels ones.
Moreover, we change the number of decoder layers. As shown in Table~\ref{table: comparison Transformer layers number}, the performance grows at the first three layers and saturates at the fourth decoder layer. 

\begin{table}[t]
    \small
    \centering
	\begin{tabular}{ c | c |c|cc|cc}
	C & GFLOPs & AP$^{kp}$ & AP$^{kp}_{50}$ & AP$^{kp}_{75}$ & AP$^{kp}_{M}$ & AP$^{kp}_{L}$ \\
	\hline
	128 & 2.28 & 63.2 & 85.0 & 69.8 & 60.6 & 69.8 \\
	256 & 3.61 & \textbf{64.5} & \textbf{85.2} & \textbf{71.2} & \textbf{61.5} & \textbf{71.5} \\
	\end{tabular}
	\vspace{0.5em}
	\caption{The effect of the number of channels of the input features to the Transformer encoder on COCO \emph{val2017} set. $C$ is the number of channels. In this experiment, we set the number of transformer encoder layers: $N_E=0$, and transformer decoder layers: $N_D=6$.}
	\label{table: comparison Transformer chennal}
\end{table}

\begin{table}[t]
    \small
    \centering
	\begin{tabular}{ c | c |c|cc|cc}
	$N_D$ & GFLOPs & AP$^{kp}$ & AP$^{kp}_{50}$ & AP$^{kp}_{75}$ & AP$^{kp}_{M}$ & AP$^{kp}_{L}$ \\
	\hline
	1 & 6.32 & 65.7 & 86.3 & 73.4 & 63.0 & 72.4 \\
	2 & 6.41 & 66.9 & 86.5 & 74.0 & 64.2 & 73.8 \\
	3 & 6.50 & 67.1 & 86.6 &74.2 & 64.5 & 73.9 \\
	4 & 6.59 & 67.2 & 86.6 &74.2 & 64.6 & 74.0 \\
	5 & 6.68 & 67.2 & 86.6 &74.2 & 64.6 & 74.1 \\
	6 & 6.77 & 67.2 & 86.6 &74.2 & 64.6 & 74.1 \\
	\end{tabular}
	\vspace{0.5em}
	\caption{Ablation study of different numbers of decoder layers on COCO \emph{val2017} set. $N_D$ is the number of decoder layers used for refining the location of key-points. "GFLOPs" indicates the computational cost. In this experiment, we set the number of transformer encoder layers: $N_E=6$.}
	\label{table: comparison Transformer layers number}
\end{table}

\begin{figure*}[h]
\centering 
\includegraphics[width=0.745\textwidth]{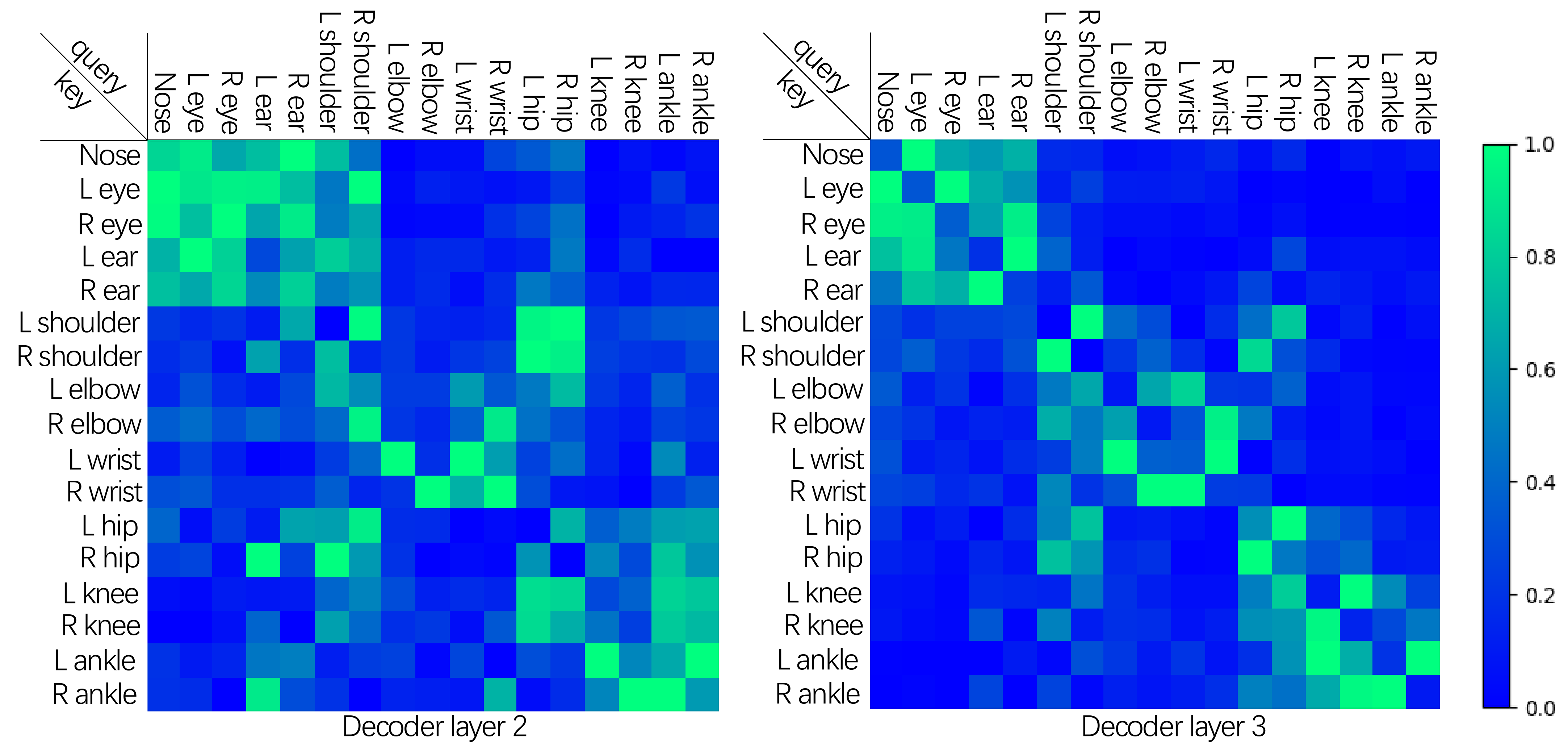}
\caption{\textbf{Visualization of the attention weights of the q2q attention}. We average the attention maps over the whole COCO \texttt{2017val} dataset. The left map is the attention weights of the second decoder layer. The right map is the attention weights of the third decoder layer. `L' means the joints are in the left. `R' means the joints are in the right. The horizontal axis and the vertical axis represent the input query and key of the attention module, respectively. Multi-head attention computes the attention weights between each pair of the queries and keys. The query attends more to the key with a higher attention weight.}
\label{fig:q2q attn}
\end{figure*}

\noindent\textbf{Auxiliary loss.}
As shown in previous works\cite{dosovitskiy2020image, zhu2020deformable, zhu2020deformable}, the transformer modules may converge slower. To mitigate this issue, we adopt the deformable attention module proposed in \cite{zhu2020deformable}. Apart from that, we propose an auxiliary loss to accelerate the convergence speed of TFPose. Here, we investigate the effect of the auxiliary loss. In this experiment, the first model is only supervised by regression loss; the second model is supervised by both regression loss and auxiliary loss. As shown in Figure~\ref{fig:aux loss} and  Table~\ref{table: comparison auxiliary loss}, the auxiliary loss can significantly accelerates the convergence speed of TFPose and boost the performance by a large margin ($+2.3\%$ AP).

\begin{table}[t]
    \small
    \centering
	\begin{tabular}{ c | c |c|cc|cc}
	aux & GFLOPs & AP$^{kp}$ & AP$^{kp}_{50}$ & AP$^{kp}_{75}$ & AP$^{kp}_{M}$ & AP$^{kp}_{L}$ \\ 
	\hline
	& 6.76& 67.2 & 86.6 &74.2 & 64.6 & 74.1 \\
	\checkmark & 6.76 & \textbf{69.5}  & \textbf{87.5} & \textbf{76.5} & \textbf{66.1} & \textbf{77.0} \\
	\end{tabular}
	\vspace{0.5em}
	\caption{Ablation study of effectiveness of auxiliary loss on COCO \emph{val2017} set. "aux" indicts whether using auxiliary loss. In this experiment, we set the number of transformer encoder layers: $N_E=6$, and transformer decoder layers: $N_D=6$.}
	\label{table: comparison auxiliary loss}
\end{table}
\vspace{-0.5em}

\subsection{Discussions on TFPose}
\begin{figure*}[h]
\centering 
\includegraphics[width=0.9\textwidth]{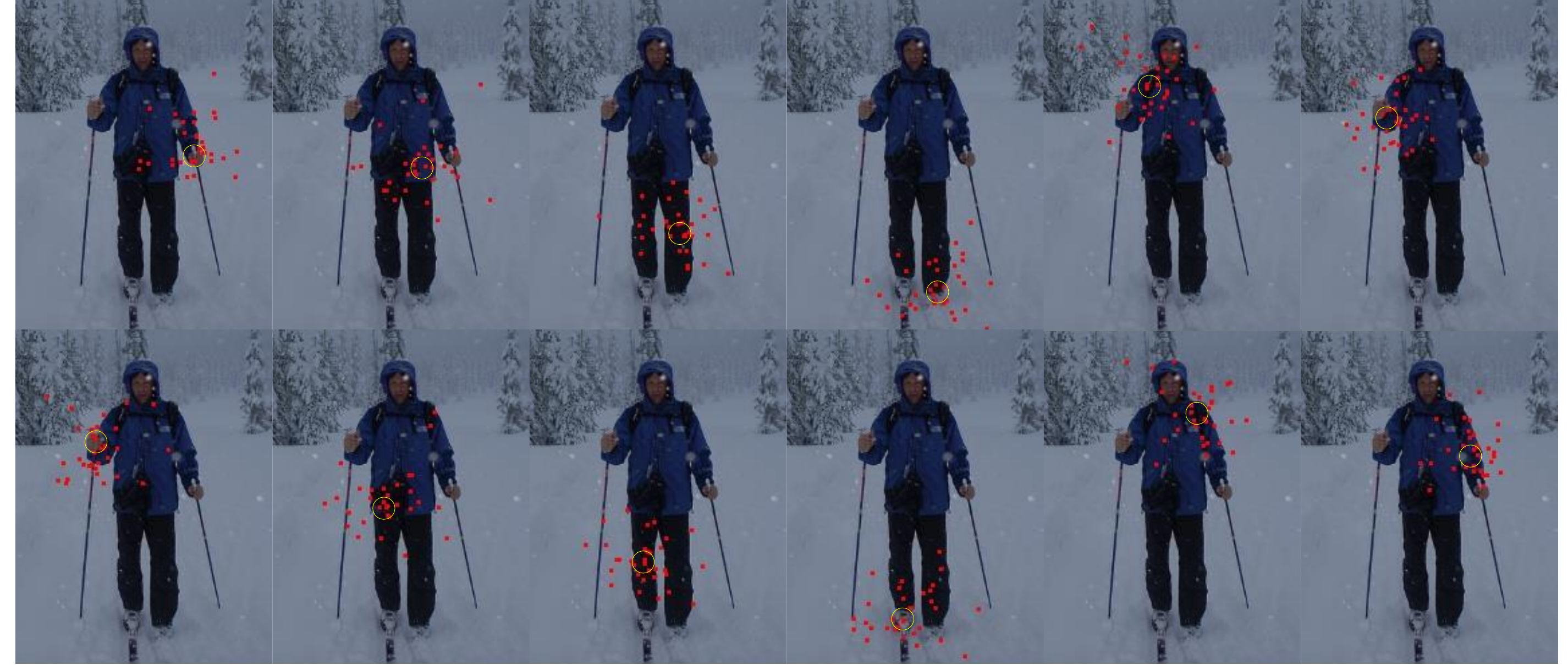}
\caption{%
\textbf{Visualisation of the sampling point on feature map}. There are 17 queries for 17 keypoints. We visualize 12 body joints queries (not including facial joints). Each image correspond to a body joints. Red dot represent the sampling point; yellow circle represent the ground truth.}
\label{fig:vis_sampling}
\end{figure*}

\noindent\textbf{Visualization of sampling keypoints.}
To study how the Deformable Attention Module locate the body joints, we visualize the sampling locations of the module on the feature maps $C_3$ . In Deformable Attention Module, there are 8 attention heads and every head will sample 4 points on every feature map. So for the $C_3$ feature map, there are 32 sampling points. As shown in Figure~\ref{fig:vis_sampling}, the sampling points (red dot) are all densely located nearby the ground truth (yellow circle). This visualization shows that TFPose can address the feature mis-alignment issue in a sense, and supervises the CNN with dense pixel information.  

\noindent\textbf{Visualization of query-to-query attention.}
To further study how the query-to-query self-attention module works, we visualize the attention weights of the query-to-query self-attention. As shown in Figure~\ref{fig:q2q attn},  there are two obvious patterns of attention: the first attention pattern is that the symmetric joints (\eg left shoulder and right shoulder) are more likely to attend to each other, and the second attention pattern is that the adjacent joints (\eg eyes, nose, and mouth) are more likely to attend to each other.

To have a better understanding of this attention pattern, we also visualize the attention graph between each keypoint according to the attention maps in the supplementary. This attention pattern suggests that TFPose can employ the context and structured relationship between the body joints to locate and classify the types of body joints.

\subsection{Comparison with State-of-the-art Methods}
\begin{table*}[t!]
    \small
    \centering
    \setlength{\tabcolsep}{12pt}
	\begin{tabular}{ r |l|c|c|c}
	    Models              & Backbone      & Input Size & GFLOPs  & AP(OKS)   \\ %
        \hline
	    DeepPose~\cite{toshev2014DeepPose}  & ResNet-101    & $256\times192$    &   7.6     & 56.0  \\
	    DeepPose~\cite{toshev2014DeepPose}  & ResNet-152    & $256\times192$    &   11.3     & 58.3  \\
	    8-stage Hourglass~\cite{newell2016stacked}   & -             & $256\times192$    &   19.5     & 66.9  \\
	    8-stage Hourglass~\cite{newell2016stacked}   & -             & $256\times256$    &   25.9     & 67.1  \\
	    CPN~\cite{chen2018cascaded}        & ResNet-50     & $256\times192$    &   6.2     & 68.6 (69.4)  \\
	    CPN~\cite{chen2018cascaded}  & ResNet-50     & $384\times288$    &   13.9     & 70.6 (71.6) \\
	    SimpleBaseline~\cite{Xiao_2018_ECCV}     & ResNet-50     & $256\times192$    &   8.9     & 70.4  \\
	    SimpleBaseline~\cite{Xiao_2018_ECCV}     & ResNet-50     & $384\times288$    &   20.0     & 72.2  \\
	    Ours ($N_D=4$)  & ResNet-50     & $256\times192$    &   7.7     & 70.5  \\
	    Ours ($N_D=6$)  & ResNet-50     & $256\times192$    &   9.2     & 71.0  \\
	    Ours ($N_D=6$)  & ResNet-50     & $384\times288$    &   20.4     & 72.4  \\
	\end{tabular}
	\vspace{0.5em}
	\caption{Comparisons with previous works on the COCO \emph{val} split. For CPN, the results in the brackets are with online hard keypoints mining.
	All the reported methods use person detectors with similar performance. Specifically, Hourglass and CPN use the person detector with 55.3\% AP on COCO. Others use the person detector with 56.4\% AP.
	DeepPose is implemented by the mmpose~\cite{mmpose2020}. Flipping test is applied for all model. $N_D$ represents the number of encoder layers.}
	\label{table:comparisons_on_val2017}
\end{table*}

In this section, we compare TFPose with previous state-of-the-art 2D pose estimation methods on COCO \texttt{val} split, COCO \texttt{test}-\texttt{dev} split and MPII\cite{mpii}. We compare these method in terms of both accuracy and computational cost. The results of our proposed TFPose and %
other state-of-the-art methods
are listed in Table~\ref{table:comparisons_on_val2017}, Table~\ref{table:comparisons_with_sota} and Table~\ref{table:comparisons_on_mpii}.

\begin{table*}[t!]
    \small
    \centering
    \setlength{\tabcolsep}{12pt}
	\begin{tabular}{p{2.7cm}<{\raggedleft}|p{1.8cm}|c|c|c|p{0.45cm}<{\centering}p{0.45cm}<{\centering}|p{0.45cm}<{\centering}p{0.45cm}<{\centering}}
	    Method                                      & Backbone  & Input Size   & GFLOPs     & AP$^{kp}$ & AP$^{kp}_{50}$ & AP$^{kp}_{75}$ & AP$^{kp}_{M}$ & AP$^{kp}_{L}$ \\
	    \hline
	    \multicolumn{8}{c}{\textbf{Heatmap-based methods}} \\
	    \hline
	    AE~\cite{newell2016associative}             & HourGlass~\cite{newell2016stacked}&        -        &         -      & 56.6 & 81.8 & 61.8 & 49.8 & 67.0 \\
	    Mask R-CNN~\cite{he2017mask}                & ResNet-50                         &        -        &     -          & 62.7 & 87.0 & 68.4 & 57.4 & 71.1 \\
	    CMU-Pose~\cite{cao2019openpose}             & VGG-19~\cite{Simonyan15}          &        -        &      -         & 64.2 & 86.2 & 70.1 & 61.0 & 68.8 \\
        G-RMI~\cite{papandreou2017towards}          & ResNet-101                        &        -        &      -         & 64.9 & 85.5 & 71.3 & 62.3 & 70.0 \\
        HigherHRNet$^{\dag\ddagger}$~\cite{cheng2020higherhrnet} & HRNet-W48            &      -          &       -        & 70.5 & 89.3 & 77.2 & 66.6 & 75.8 \\
        CPN~\cite{chen2018cascaded}                 & ResNet-Ince.                  & $384\times288$ &  29.2         & 72.1 & 91.4 & 80.0 & 68.7 & 77.2 \\
        SimpleBaseline$^\dag$~\cite{xiao2018simple} & ResNet-50                        & $256\times192$ &  8.9         & 70.0 & 90.9 & 77.9 & 66.8 & 75.8 \\
        SimpleBaseline$^\dag$~\cite{xiao2018simple} & ResNet-152                        & $384\times288$ &  35.6         & 73.7 & 91.9 & 81.1 & 70.3 & 80.0 \\
        HRNet$^\dag$~\cite{sun2019deep}             & HRNet-W32                         & $384\times288$ &  16.0         & 74.9 & 92.5 & 82.8 & 71.3 & 80.9 \\
        \hline
        \multicolumn{8}{c}{\textbf{Regression-based methods}} \\
	    \hline
	    DeepPose$^\dag$~\cite{toshev2014DeepPose}           & ResNet-101                        & $256\times192$ &  7.69         & 57.4 & 86.5 & 64.2 & 55.0 & 62.8 \\
	    DeepPose$^\dag$~\cite{toshev2014DeepPose}             & ResNet-152                        & $256\times192$ &  11.34        & 59.3 & 87.6 & 66.7 & 56.8 & 64.9 \\
	    Directpose~\cite{tian2019directpose}            & ResNet-50                         &        -        &   -            & 62.2 & 86.4 & 68.2 & 56.7 & 69.8 \\
	    Directpose~\cite{tian2019directpose}            & ResNet-101                        &        -        &   -            & 63.3 & 86.7 & 69.4 & 57.8 & 71.2 \\
	    SPM~\cite{nie2019single}                         & HourGlass~\cite{newell2016stacked}&        -        &  -             & 66.9 & 88.5 & 72.9 & 62.6 & 73.1 \\
		Int. Reg.~\cite{sun2018integral}            & ResNet-101                        & $256\times256$ &     11.0      & 67.8 & 88.2 & 74.8 & 63.9 & 74.0 \\
		Ours$^\dag$($N_D =4$)                   & ResNet-50                         & $256\times192$ &   7.7      & 70.5 & 90.4 & 78.7 & 67.6 & 76.8 \\
		Ours$^\dag$($N_D=6$)                   & ResNet-50                         & $256\times192$ &   9.2      & 70.9 & 90.5 & 79.0 & 68.1 & 77.0 \\
		Ours$^\dag$($N_D=6$)                    & ResNet-50                         & $384\times288$ &   20.4      &72.2  & 90.9 &80.1  &69.1  & 78.8 \\
	\end{tabular}
	\vspace{0.5em}
	\caption{Comparisons with state-of-the-art methods on COCO \emph{test-dev} set. $^\dag$ and $^\ddagger$ denote flipping and multi-sacle testing, respectively. Input size and the GFLOPs are shown for the single person pose estimation methods. 'ResNet-Ince.' represent the ResNet inception. The Simple baseline(ResNet-50) is tested with the official code. $N_D$ represents the number of encoder layers.
	}
	\label{table:comparisons_with_sota}
\end{table*}

\begin{table*}[t!]
    \small
    \centering
    \setlength{\tabcolsep}{12pt}
	\begin{tabular}{ r  |c|c|c|c|c|c|c|c}
	Method & Head & Sho. & Elb. & Wri. & Hip & Knee & Ank. & Total \\
    \hline
    \multicolumn{8}{c}{\textbf{Heatmap-based methods}} \\
    \hline
    Pishchulin et al. \cite{pishchulin2013strong}       & 74.3 & 49.0 & 40.8 & 34.1 & 36.5 & 34.4 & 35.2 & 44.1 \\
    Tompson et al. \cite{tompson2014joint}              & 95.8 & 90.3 & 80.5 & 74.3 & 77.6 & 69.7 & 62.8 & 79.6 \\
    Hu et al. \cite{hu2016bottom}                       & 95.0 & 91.6 & 83.0 & 76.6 & 81.9 & 74.5 & 69.5 & 82.4 \\
    Lifshitz et al. \cite{lifshitz2016human}            & 97.8 & 93.3 & 85.7 & 80.4 & 85.3 & 76.6 & 70.2 & 85.0 \\
    Raf et al. \cite{rafi2016efficient}                 & 97.2 & 93.9 & 86.4 & 81.3 & 86.8 & 80.6 & 73.4 & 86.3 \\
    Bulat et al. \cite{bulat2016human}                  & 97.9 & 95.1 & 89.9 & 85.3 & 89.4 & 85.7 & 81.7 & 89.7 \\
    Chu et al. \cite{chu2017multi}                      & 98.5 & 96.3 & 91.9 & 88.1 & 90.6 & 88.0 & 85.0 & 91.5 \\
    Ke et al. \cite{Ke_2018_ECCV}                       & 98.5 & 96.8 & 92.7 & 88.4 & 90.6 & 89.3 & 86.3 & 92.1 \\
    Tang et al. \cite{Tang_2018_ECCV}                  & 98.4 & 96.9 & 92.6 & 88.7 & 91.8 & 89.4 & 86.2 & 92.3 \\  
    Zhang et al. \cite{zhang2019human}                  & 98.6 & 97.0 & 92.8 & 88.8 & 91.7 & 89.8 & 86.6 & 92.5 \\    
    \hline
    \multicolumn{8}{c}{\textbf{Regression-based methods}} \\
    \hline
    Carreira et al. \cite{carreira2016human}      & 95.7 & 91.7 & 81.7 & 72.4 & 82.8 & 73.2 & 66.4 & 81.3 \\
    Sun et al. \cite{sun2017compositional}             & 97.5 & 94.3 & 87.0 & 81.2 & 86.5 & 78.5 & 75.4 & 86.4 \\
    Aiden et al. (ResNet-50)\cite{nibali2018numerical} & 97.8 & 96.0 & 90.0 & 84.3 & 89.8 & 85.2 & 79.7 & 89.5 \\ 
    Ours  (ResNet-50)                         & 98.0 & 95.9 & 91.0 & 86.0 & 89.8 & 86.6 & 82.6 & 90.4 \\ 
	\end{tabular}
	\vspace{0.5em}
	\caption{MPII human pose \emph{test} set PCKh accuracies. For our model, the number of encoder layers $N_D$ is set to 6.}
	\label{table:comparisons_on_mpii}
\end{table*}

\noindent\textbf{Results on COCO \emph{val.}\  set.} As shown in Table~\ref{table:comparisons_on_val2017}, with similar computational cost, TFPose with 4 encoder layers and ResNet-50 surpass the previous regression-based method DeepPose with ResNet-101 (70.5\% AP vs. 56.0\% AP) by a large margin and even has much better performance than DeepPose with ResNet-152 (70.5\% AP vs.\  58.3\% AP). Besides, TFPose also outperform many heatmap-based methods, for example, 8-stage Hourglass\cite{newell2016stacked}(70.5\% AP vs. 67.1\% AP), CPN\cite{chen2018cascaded}(70.5\% AP vs. 69.4\% AP) by a large margin. It is also important to note that TFPose with 4 encoder layers and ResNet-50 outperforms the strong baseline SimpleBaseline\cite{Xiao_2018_ECCV} with ResNet-50 (70.5\% AP vs. 70.4\% AP) with lower computational cost (7.68 GFLOPs vs. 8.9 GFLOPs).

\noindent\textbf{Results on COCO \emph{test-dev} set.}
As shown in Table~\ref{table:comparisons_with_sota}, TFPose achieves the best result among regression-based methods. Especially, TFPose with 6 encoder layers and ResNet-50 achieves 70.9\% AP, which is higher than the Int. Reg.\cite{sun2018integral} (67.8\% AP), and our computational cost is lower than the Int. Reg. (9.15 GFLOPs vs. 11.0 GFLOPs). Moreover, with the same bacbone ResNet-50, our TFPose even achieves better performance than the strong heatmap-based method SimpleBaseline (70.5\% vs. 70.0\% AP) with less computational complexity (7.7 GFLOPS vs. 8.9 GFLOPS). Additionally, the results of TFPose are also close to the best reported pose estimation results. For exmaple, the performance of TFPose (72.2\% AP) is close to the ResNet-Inception based CPN(72.1\% AP) and ResNet-152 based SimpleBaseline (73.7\% AP). Note that they use much larger backbones than ours. %

\noindent\textbf{Results on MPII \emph{test} set.} 
On the MPII benchmark, TFPose also achieves the best results among the regression-based methods. As shown in Table~\ref{table:comparisons_on_mpii}, TFPose with ResNet-50 is higher than the method proposed by Aiden et al.\cite{nibali2018numerical} (90.4\% vs. 89.5\%) with the same backbone. TFPose is also comparable to heatmap-based methods. 
\section{Conclusion}

We have proposed a novel pose estimation framework named TFPose built upon Transformers, which largely improves the performance of the regression-based pose estimation and bypasses the drawbacks of heatmap-based methods such as the non-differentiable post-processing and quantization error. We have shown that with the attention mechanism, TFPose can naturally capture the structured relationship between the body joints, resulting in improved performance. Extensive experiments on the MS-COCO and MPII benchmarks show that TFPose can achieve state-of-the-art performance among regression-based methods and is comparable to the best heatmap-based methods.

\section*{Acknowledgements}
The  authors would  like to  thank Alibaba Group
for the donation of GPU cloud computing resources.

{\small
\bibliographystyle{ieee_fullname}
\bibliography{tfpose}

\begin{thebibliography}{10}\itemsep=-1pt

\bibitem{mpii}
Mykhaylo Andriluka, Leonid Pishchulin, Peter Gehler, and Bernt Schiele.
\newblock 2d human pose estimation: New benchmark and state of the art
  analysis.
\newblock In {\em {Proc. IEEE Conf. Comput. Vis. Pattern Recog.}}, pages
  3686--3693, 2014.

\bibitem{brown2020language}
Tom~B Brown, Benjamin Mann, Nick Ryder, Melanie Subbiah, Jared Kaplan, Prafulla
  Dhariwal, Arvind Neelakantan, Pranav Shyam, Girish Sastry, Amanda Askell,
  et~al.
\newblock Language models are few-shot learners.
\newblock In {\em {Proc. Adv. Neural Inform. Process. Syst.}}, 2020.

\bibitem{bulat2016human}
Adrian Bulat and Georgios Tzimiropoulos.
\newblock Human pose estimation via convolutional part heatmap regression.
\newblock In {\em {Proc. Eur. Conf. Comput. Vis.}}, pages 717--732, 2016.

\bibitem{cai2020learning}
Yuanhao Cai, Zhicheng Wang, Zhengxiong Luo, Binyi Yin, Angang Du, Haoqian Wang,
  Xiangyu Zhang, Xinyu Zhou, Erjin Zhou, and Jian Sun.
\newblock Learning delicate local representations for multi-person pose
  estimation.
\newblock In {\em {Proc. Eur. Conf. Comput. Vis.}}, pages 455--472. Springer,
  2020.

\bibitem{cao2019openpose}
Zhe Cao, Gines Hidalgo, Tomas Simon, Shih-En Wei, and Yaser Sheikh.
\newblock Openpose: realtime multi-person 2d pose estimation using part
  affinity fields.
\newblock {\em {IEEE Trans. Pattern Anal. Mach. Intell.}}, 43(1):172--186,
  2019.

\bibitem{carion2020end}
Nicolas Carion, Francisco Massa, Gabriel Synnaeve, Nicolas Usunier, Alexander
  Kirillov, and Sergey Zagoruyko.
\newblock End-to-end object detection with transformers.
\newblock In {\em {Proc. Eur. Conf. Comput. Vis.}}, pages 213--229. Springer,
  2020.

\bibitem{carreira2016human}
Joao Carreira, Pulkit Agrawal, Katerina Fragkiadaki, and Jitendra Malik.
\newblock Human pose estimation with iterative error feedback.
\newblock In {\em {Proc. IEEE Conf. Comput. Vis. Pattern Recog.}}, pages
  4733--4742, 2016.

\bibitem{chen2020pre}
Hanting Chen, Yunhe Wang, Tianyu Guo, Chang Xu, Yiping Deng, Zhenhua Liu, Siwei
  Ma, Chunjing Xu, Chao Xu, and Wen Gao.
\newblock Pre-trained image processing transformer.
\newblock {\em arXiv preprint arXiv:2012.00364}, 2020.

\bibitem{chen2018cascaded}
Yilun Chen, Zhicheng Wang, Yuxiang Peng, Zhiqiang Zhang, Gang Yu, and Jian Sun.
\newblock Cascaded pyramid network for multi-person pose estimation.
\newblock In {\em {Proc. IEEE Conf. Comput. Vis. Pattern Recog.}}, pages
  7103--7112, 2018.

\bibitem{cheng2020higherhrnet}
Bowen Cheng, Bin Xiao, Jingdong Wang, Honghui Shi, Thomas~S Huang, and Lei
  Zhang.
\newblock Higherhrnet: Scale-aware representation learning for bottom-up human
  pose estimation.
\newblock In {\em {Proc. IEEE Conf. Comput. Vis. Pattern Recog.}}, pages
  5386--5395, 2020.

\bibitem{chu2017multi}
Xiao Chu, Wei Yang, Wanli Ouyang, Cheng Ma, Alan~L Yuille, and Xiaogang Wang.
\newblock Multi-context attention for human pose estimation.
\newblock In {\em {Proc. IEEE Conf. Comput. Vis. Pattern Recog.}}, pages
  1831--1840, 2017.

\bibitem{mmpose2020}
MMPose Contributors.
\newblock Openmmlab pose estimation toolbox and benchmark.
\newblock \url{https://github.com/open-mmlab/mmpose}, 2020.

\bibitem{deng2009imagenet}
Jia Deng, Wei Dong, Richard Socher, Li-Jia Li, Kai Li, and Li Fei-Fei.
\newblock Imagenet: A large-scale hierarchical image database.
\newblock In {\em {Proc. IEEE Conf. Comput. Vis. Pattern Recog.}}, pages
  248--255. Ieee, 2009.

\bibitem{devlin2018bert}
Jacob Devlin, Ming-Wei Chang, Kenton Lee, and Kristina Toutanova.
\newblock Bert: Pre-training of deep bidirectional transformers for language
  understanding.
\newblock {\em arXiv preprint arXiv:1810.04805}, 2018.

\bibitem{dosovitskiy2020image}
Alexey Dosovitskiy, Lucas Beyer, Alexander Kolesnikov, Dirk Weissenborn,
  Xiaohua Zhai, Thomas Unterthiner, Mostafa Dehghani, Matthias Minderer, Georg
  Heigold, Sylvain Gelly, et~al.
\newblock An image is worth 16x16 words: Transformers for image recognition at
  scale.
\newblock {\em arXiv preprint arXiv:2010.11929}, 2020.

\bibitem{han2020survey}
Kai Han, Yunhe Wang, Hanting Chen, Xinghao Chen, Jianyuan Guo, Zhenhua Liu,
  Yehui Tang, An Xiao, Chunjing Xu, Yixing Xu, et~al.
\newblock A survey on visual transformer.
\newblock {\em arXiv preprint arXiv:2012.12556}, 2020.

\bibitem{he2017mask}
Kaiming He, Georgia Gkioxari, Piotr Doll{\'a}r, and Ross Girshick.
\newblock Mask r-cnn.
\newblock In {\em {Proc. Int. Conf. Comput. Vis.}}, pages 2961--2969, 2017.

\bibitem{he2016deep}
Kaiming He, Xiangyu Zhang, Shaoqing Ren, and Jian Sun.
\newblock Deep residual learning for image recognition.
\newblock In {\em {Proc. IEEE Conf. Comput. Vis. Pattern Recog.}}, pages
  770--778, 2016.

\bibitem{hu2016bottom}
Peiyun Hu and Deva Ramanan.
\newblock Bottom-up and top-down reasoning with hierarchical rectified
  gaussians.
\newblock In {\em {Proc. IEEE Conf. Comput. Vis. Pattern Recog.}}, pages
  5600--5609, 2016.

\bibitem{hu2018facial}
Tao Hu, Honggang Qi, Jizheng Xu, and Qingming Huang.
\newblock Facial landmarks detection by self-iterative regression based
  landmarks-attention network.
\newblock volume~32, 2018.

\bibitem{huang2020devil}
Junjie Huang, Zheng Zhu, Feng Guo, and Guan Huang.
\newblock The devil is in the details: Delving into unbiased data processing
  for human pose estimation.
\newblock In {\em {Proc. IEEE Conf. Comput. Vis. Pattern Recog.}}, pages
  5700--5709, 2020.

\bibitem{huang2020hand}
Lin Huang, Jianchao Tan, Ji Liu, and Junsong Yuan.
\newblock Hand-transformer: Non-autoregressive structured modeling for 3d hand
  pose estimation.
\newblock In {\em {Proc. Eur. Conf. Comput. Vis.}}, pages 17--33. Springer,
  2020.

\bibitem{huang2020hot}
Lin Huang, Jianchao Tan, Jingjing Meng, Ji Liu, and Junsong Yuan.
\newblock Hot-net: Non-autoregressive transformer for 3d hand-object pose
  estimation.
\newblock In {\em {Proc. ACM Int. Conf. Multimedia}}, pages 3136--3145, 2020.

\bibitem{Ke_2018_ECCV}
Lipeng Ke, Ming-Ching Chang, Honggang Qi, and Siwei Lyu.
\newblock Multi-scale structure-aware network for human pose estimation.
\newblock In {\em {Proc. Eur. Conf. Comput. Vis.}}, September 2018.

\bibitem{li2019crowdpose}
Jiefeng Li, Can Wang, Hao Zhu, Yihuan Mao, Hao-Shu Fang, and Cewu Lu.
\newblock Crowdpose: Efficient crowded scenes pose estimation and a new
  benchmark.
\newblock In {\em {Proc. IEEE Conf. Comput. Vis. Pattern Recog.}}, pages
  10863--10872, 2019.

\bibitem{li2019rethinking}
Wenbo Li, Zhicheng Wang, Binyi Yin, Qixiang Peng, Yuming Du, Tianzi Xiao, Gang
  Yu, Hongtao Lu, Yichen Wei, and Jian Sun.
\newblock Rethinking on multi-stage networks for human pose estimation.
\newblock {\em arXiv preprint arXiv:1901.00148}, 2019.

\bibitem{lifshitz2016human}
Ita Lifshitz, Ethan Fetaya, and Shimon Ullman.
\newblock Human pose estimation using deep consensus voting.
\newblock In {\em {Proc. Eur. Conf. Comput. Vis.}}, pages 246--260, 2016.

\bibitem{lin2020end}
Kevin Lin, Lijuan Wang, and Zicheng Liu.
\newblock End-to-end human pose and mesh reconstruction with transformers.
\newblock {\em arXiv preprint arXiv:2012.09760}, 2020.

\bibitem{lin2014microsoft}
Tsung-Yi Lin, Michael Maire, Serge Belongie, James Hays, Pietro Perona, Deva
  Ramanan, Piotr Doll{\'a}r, and C~Lawrence Zitnick.
\newblock Microsoft coco: Common objects in context.
\newblock In {\em {Proc. Eur. Conf. Comput. Vis.}}, pages 740--755. Springer,
  2014.

\bibitem{loshchilov2017decoupled}
Ilya Loshchilov and Frank Hutter.
\newblock Decoupled weight decay regularization.
\newblock In {\em {Proc. Int. Conf. Learn. Represent.}}, 2019.

\bibitem{luo2020rethinking}
Zhengxiong Luo, Zhicheng Wang, Yan Huang, Tieniu Tan, and Erjin Zhou.
\newblock Rethinking the heatmap regression for bottom-up human pose
  estimation.
\newblock {\em arXiv preprint arXiv:2012.15175}, 2020.

\bibitem{newell2016associative}
Alejandro Newell, Zhiao Huang, and Jia Deng.
\newblock Associative embedding: End-to-end learning for joint detection and
  grouping.
\newblock {\em arXiv preprint arXiv:1611.05424}, 2016.

\bibitem{newell2016stacked}
Alejandro Newell, Kaiyu Yang, and Jia Deng.
\newblock Stacked hourglass networks for human pose estimation.
\newblock In {\em {Proc. Eur. Conf. Comput. Vis.}}, pages 483--499. Springer,
  2016.

\bibitem{nibali2018numerical}
Aiden Nibali, Zhen He, Stuart Morgan, and Luke Prendergast.
\newblock Numerical coordinate regression with convolutional neural networks.
\newblock {\em arXiv preprint arXiv:1801.07372}, 2018.

\bibitem{nie2019single}
Xuecheng Nie, Jiashi Feng, Jianfeng Zhang, and Shuicheng Yan.
\newblock Single-stage multi-person pose machines.
\newblock In {\em {Proc. Int. Conf. Comput. Vis.}}, pages 6951--6960, 2019.

\bibitem{papandreou2017towards}
George Papandreou, Tyler Zhu, Nori Kanazawa, Alexander Toshev, Jonathan
  Tompson, Chris Bregler, and Kevin Murphy.
\newblock Towards accurate multi-person pose estimation in the wild.
\newblock In {\em {Proc. IEEE Conf. Comput. Vis. Pattern Recog.}}, pages
  4903--4911, 2017.

\bibitem{pishchulin2013strong}
Leonid Pishchulin, Mykhaylo Andriluka, Peter Gehler, and Bernt Schiele.
\newblock Strong appearance and expressive spatial models for human pose
  estimation.
\newblock In {\em {Proc. Int. Conf. Comput. Vis.}}, pages 3487--3494, 2013.

\bibitem{rafi2016efficient}
Umer Rafi, Bastian Leibe, Juergen Gall, and Ilya Kostrikov.
\newblock An efficient convolutional network for human pose estimation.
\newblock In {\em {Proc. Brit. Mach. Vis. Conf.}}, volume~1, page~2, 2016.

\bibitem{Simonyan15}
Karen Simonyan and Andrew Zisserman.
\newblock Very deep convolutional networks for large-scale image recognition.
\newblock In {\em {Proc. Int. Conf. Learn. Represent.}}, 2015.

\bibitem{sun2019deep}
Ke Sun, Bin Xiao, Dong Liu, and Jingdong Wang.
\newblock Deep high-resolution representation learning for human pose
  estimation.
\newblock In {\em {Proc. IEEE Conf. Comput. Vis. Pattern Recog.}}, pages
  5693--5703, 2019.

\bibitem{sun2017compositional}
Xiao Sun, Jiaxiang Shang, Shuang Liang, and Yichen Wei.
\newblock Compositional human pose regression.
\newblock In {\em {Proc. Int. Conf. Comput. Vis.}}, pages 2602--2611, 2017.

\bibitem{sun2018integral}
Xiao Sun, Bin Xiao, Fangyin Wei, Shuang Liang, and Yichen Wei.
\newblock Integral human pose regression.
\newblock In {\em {Proc. Eur. Conf. Comput. Vis.}}, pages 529--545, 2018.

\bibitem{Tang_2018_ECCV}
Wei Tang, Pei Yu, and Ying Wu.
\newblock Deeply learned compositional models for human pose estimation.
\newblock In {\em {Proc. Eur. Conf. Comput. Vis.}}, September 2018.

\bibitem{tian2019directpose}
Zhi Tian, Hao Chen, and Chunhua Shen.
\newblock Directpose: Direct end-to-end multi-person pose estimation.
\newblock {\em arXiv preprint arXiv:1911.07451}, 2019.

\bibitem{tompson2014joint}
Jonathan Tompson, Arjun Jain, Yann LeCun, and Christoph Bregler.
\newblock Joint training of a convolutional network and a graphical model for
  human pose estimation.
\newblock {\em arXiv preprint arXiv:1406.2984}, 2014.

\bibitem{toshev2014DeepPose}
Alexander Toshev and Christian Szegedy.
\newblock Deeppose: Human pose estimation via deep neural networks.
\newblock In {\em {Proc. IEEE Conf. Comput. Vis. Pattern Recog.}}, pages
  1653--1660, 2014.

\bibitem{vaswani2017attention}
Ashish Vaswani, Noam Shazeer, Niki Parmar, Jakob Uszkoreit, Llion Jones,
  Aidan~N Gomez, Lukasz Kaiser, and Illia Polosukhin.
\newblock Attention is all you need.
\newblock In {\em {Proc. Adv. Neural Inform. Process. Syst.}}, pages
  5998--6008, 2017.

\bibitem{wang2020max}
Huiyu Wang, Yukun Zhu, Hartwig Adam, Alan Yuille, and Liang-Chieh Chen.
\newblock {MaX-DeepLab}: End-to-end panoptic segmentation with mask
  transformers.
\newblock In {\em {Proc. IEEE Conf. Comput. Vis. Pattern Recog.}}, 2021.

\bibitem{wang2020end}
Yuqing Wang, Zhaoliang Xu, Xinlong Wang, Chunhua Shen, Baoshan Cheng, Hao Shen,
  and Huaxia Xia.
\newblock End-to-end video instance segmentation with transformers.
\newblock In {\em {Proc. IEEE Conf. Comput. Vis. Pattern Recog.}}, 2021.

\bibitem{wang2018mscoco}
Zhicheng Wang, Wenbo Li, Binyi Yin, Qixiang Peng, Tianzi Xiao, Yuming Du,
  Zeming Li, Xiangyu Zhang, Gang Yu, and Jian Sun.
\newblock Mscoco keypoints challenge 2018.
\newblock In {\em {Proc. Eur. Conf. Comput. Vis.}}, volume~5, 2018.

\bibitem{xiao2018simple}
Bin Xiao, Haiping Wu, and Yichen Wei.
\newblock Simple baselines for human pose estimation and tracking.
\newblock In {\em {Proc. Eur. Conf. Comput. Vis.}}, pages 466--481, 2018.

\bibitem{Xiao_2018_ECCV}
Bin Xiao, Haiping Wu, and Yichen Wei.
\newblock Simple baselines for human pose estimation and tracking.
\newblock In {\em {Proc. Eur. Conf. Comput. Vis.}}, September 2018.

\bibitem{zhang2019human}
Hong Zhang, Hao Ouyang, Shu Liu, Xiaojuan Qi, Xiaoyong Shen, Ruigang Yang, and
  Jiaya Jia.
\newblock Human pose estimation with spatial contextual information.
\newblock {\em arXiv preprint arXiv:1901.01760}, 2019.

\bibitem{zhu2020deformable}
Xizhou Zhu, Weijie Su, Lewei Lu, Bin Li, Xiaogang Wang, and Jifeng Dai.
\newblock Deformable {DETR}: Deformable {T}ransformers for end-to-end object
  detection.
\newblock In {\em {Proc. Int. Conf. Learn. Represent.}}, 2021.

\end{thebibliography}
}

\appendix

\section{Qualitative Results of TFPose}
\begin{figure*}[h]
\centering 
\includegraphics[width=0.7\textwidth]{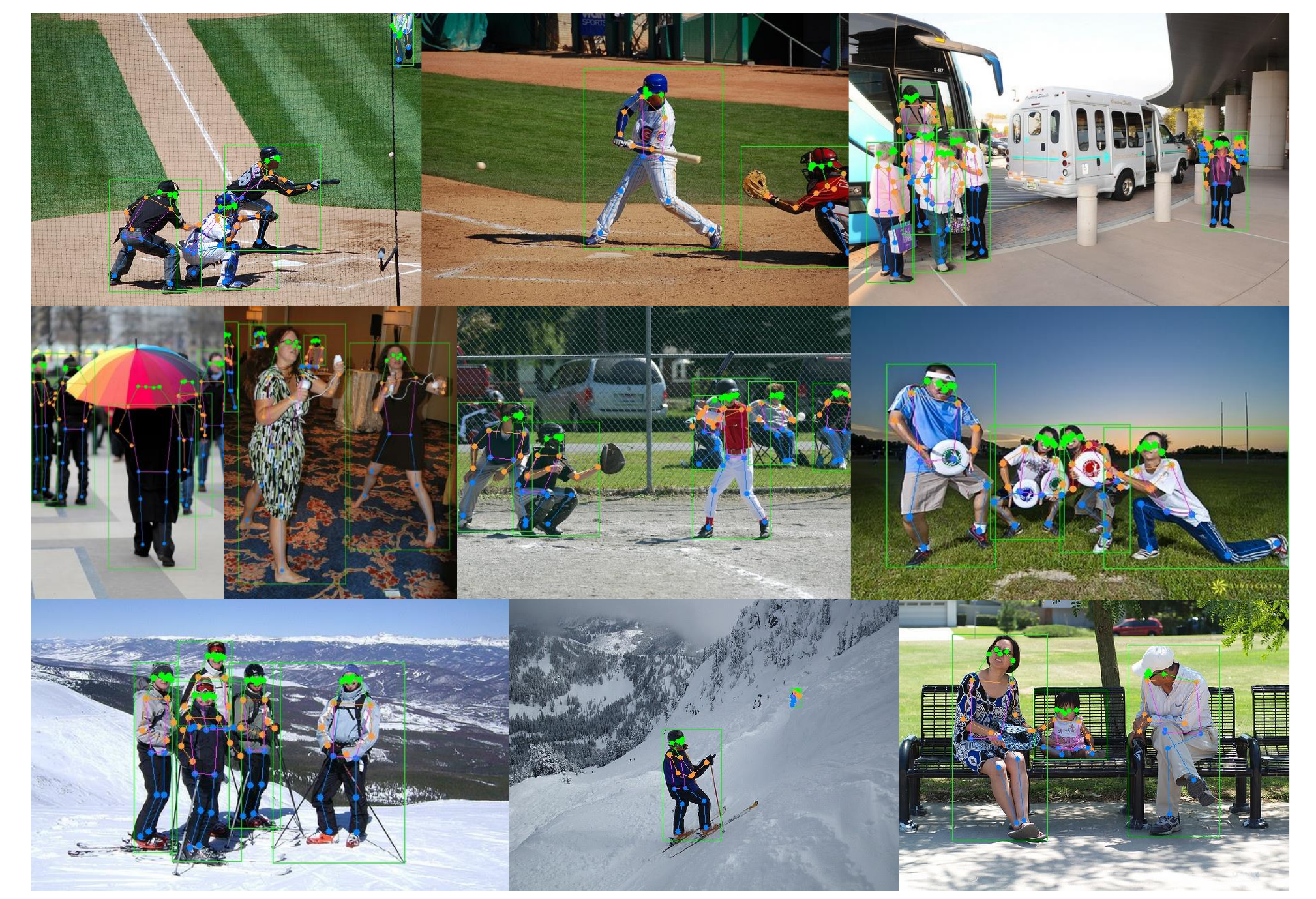}
\caption{Qualitative results of TFPose with ResNet-50 on COCO2017 \emph{val} set (single-model and singe-scale testing). The joints in upper body are represented by green and the joints in lower body are represented by blue.}
\label{fig:qualitative_results}
\end{figure*}

We show more qualitative results in Figure~\ref{fig:qualitative_results}. TFPose works reliably under various challenging cases.
\begin{figure*}[h]
\centering 
\includegraphics[width=0.680\textwidth]{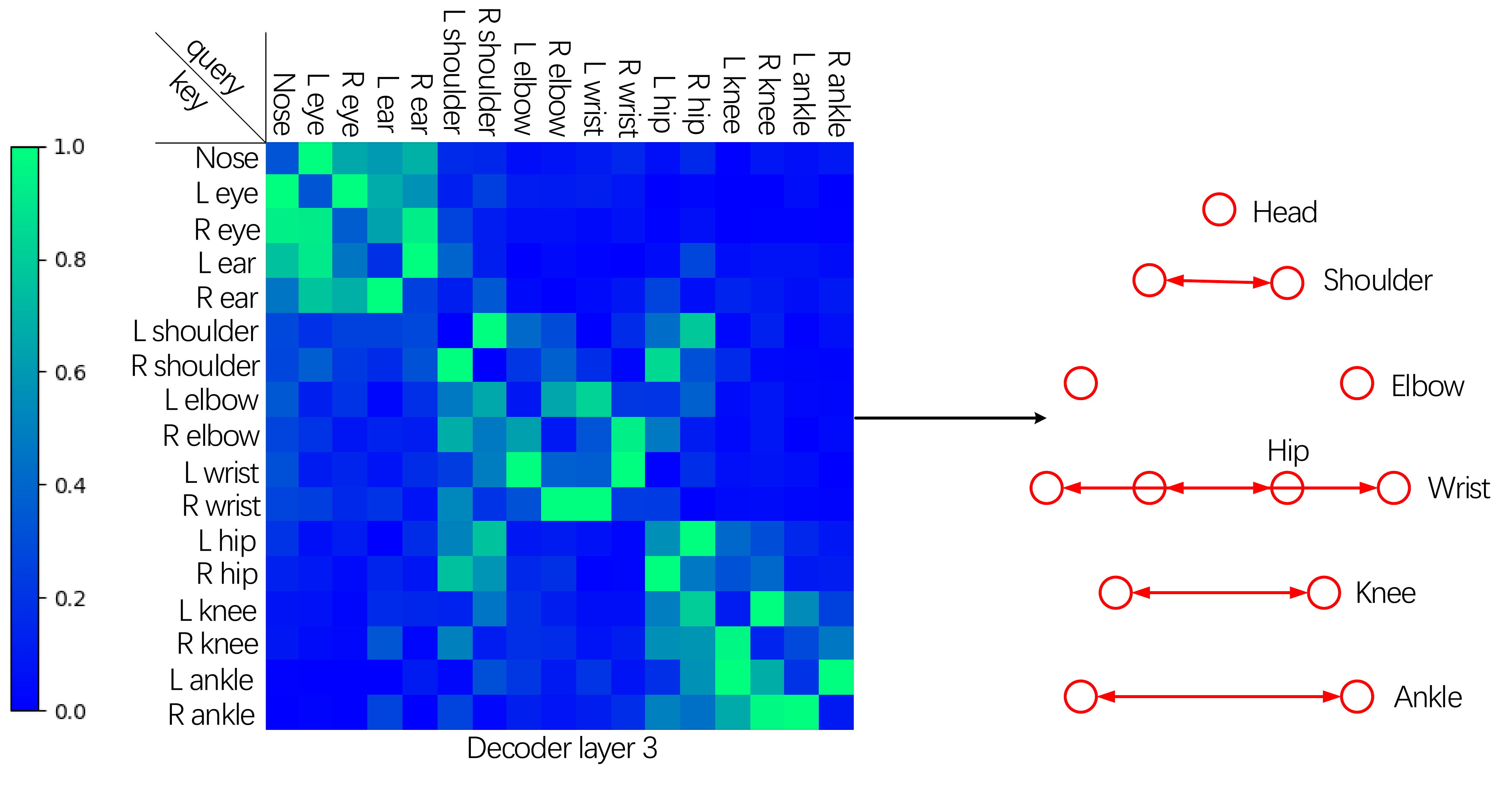}
\caption{The pattern of symmetric joints. As shown in the right graph, left shoulder and right shoulder are symmetric joints and they attend to each other.
The same pattern can be found in other body joints including left elbow and right elbow, left hip and right hip \etc.}
\label{fig:q2q_pattern_1}
\vspace{-1.5em}
\end{figure*}

\begin{figure*}[h]
\centering 
\includegraphics[width=0.680\textwidth]{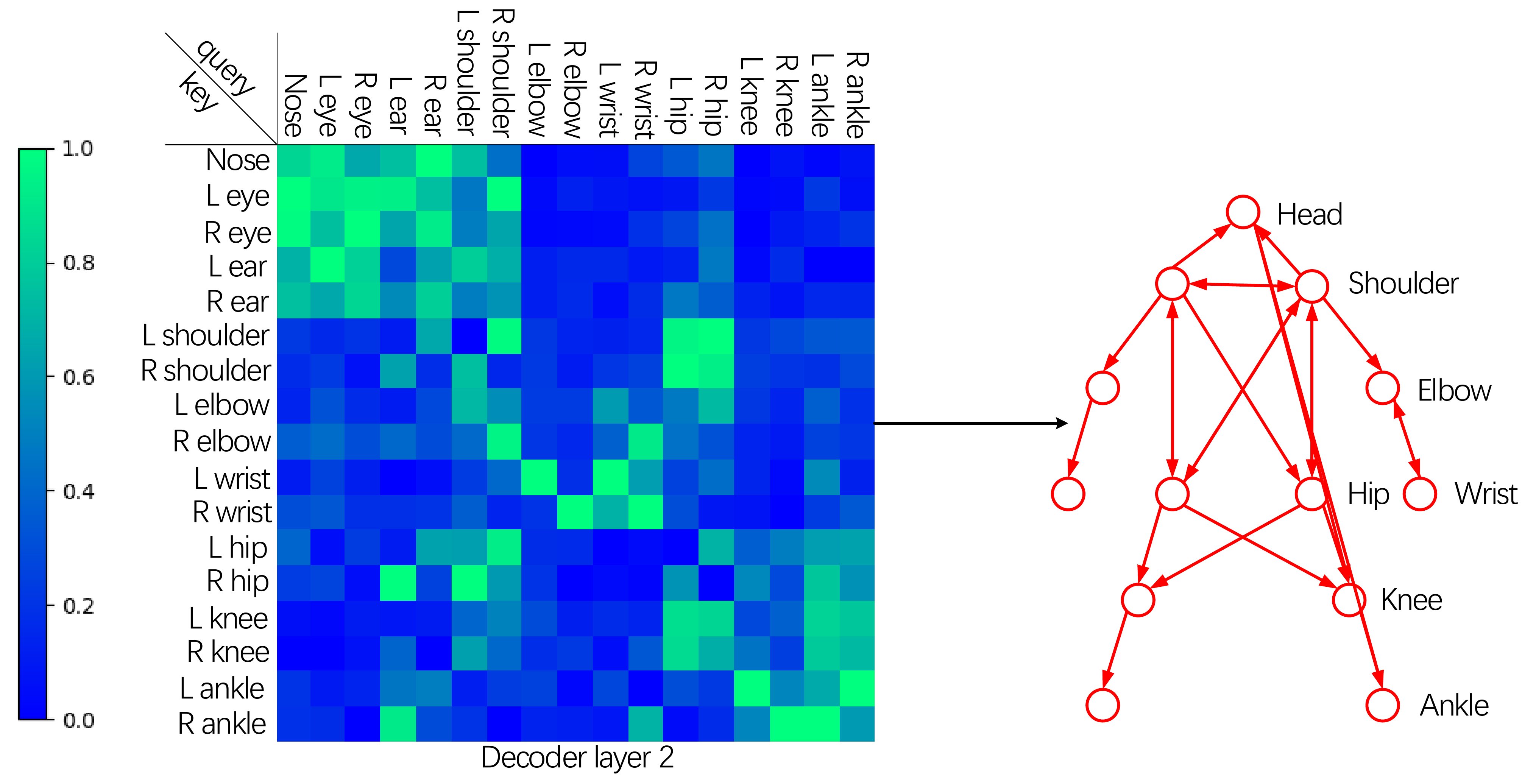}
\caption{The pattern of adjacent joints. As shown in the right graph, 
left shoulder attend to its adjacent joints including right shoulder, left elbow, and head. 
The same pattern can be found in other body joints, \eg., elbow and wrist.}
\label{fig:q2q_pattern_2}
\vspace{-1.5em}
\end{figure*}

\section{Visualization of Transformer Attentions}
\subsection{Query-to-query Attention}
We observe two obvious query-to-query attention patterns in different decoder layers, termed symmetric pattern and adjacent pattern, respectively. Both patterns exist in all decoder layers, we illustrate them separately for convenience. For symmetric pattern, Figure~\ref{fig:q2q_pattern_1} demonstrates that the correlation between all pairs of symmetric joints in the third decoder layer. For adjacent pattern, Figure~\ref{fig:q2q_pattern_2} explicitly shows that adjacent joints attend to each other in the second decoder layer.

\subsection{Multi-scale Deformable Attention}
\begin{figure*}[h]
\centering 
\includegraphics[width=0.7\textwidth]{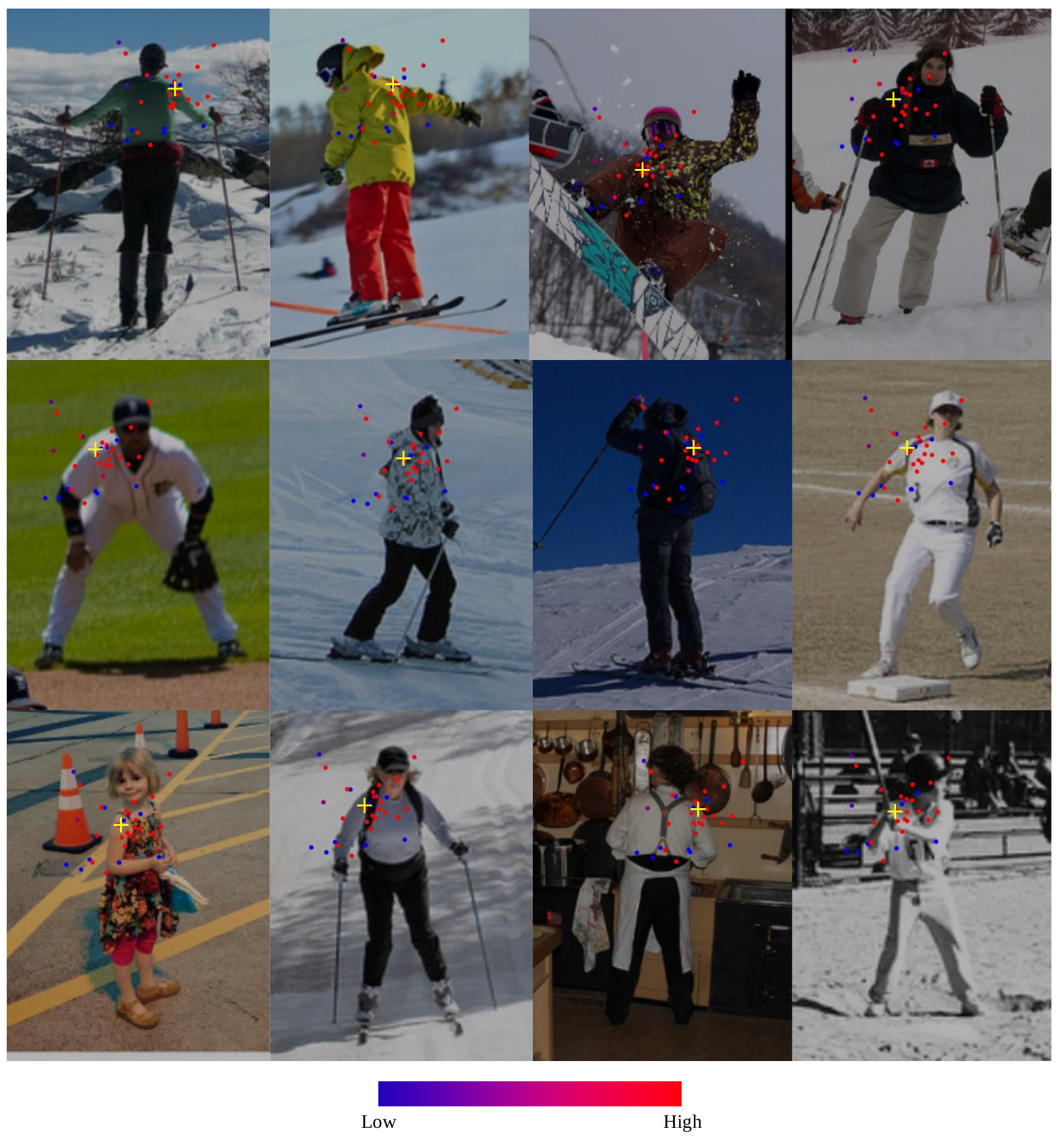}
\caption{Visualization of right shoulder's pixel-to-query attention in the last decoder layer. For readability, we draw the sampling points and attention weights from $C_3$ feature map in different pictures. Each sampling point is marked as a filled circle whose color indicates its corresponding weight. The ground truth joint is shown as yellow cross marker.}
\label{fig:right shoulder}
\vspace{-1.5em}
\end{figure*}

\begin{figure*}[h]
\centering 
\includegraphics[width=0.7\textwidth]{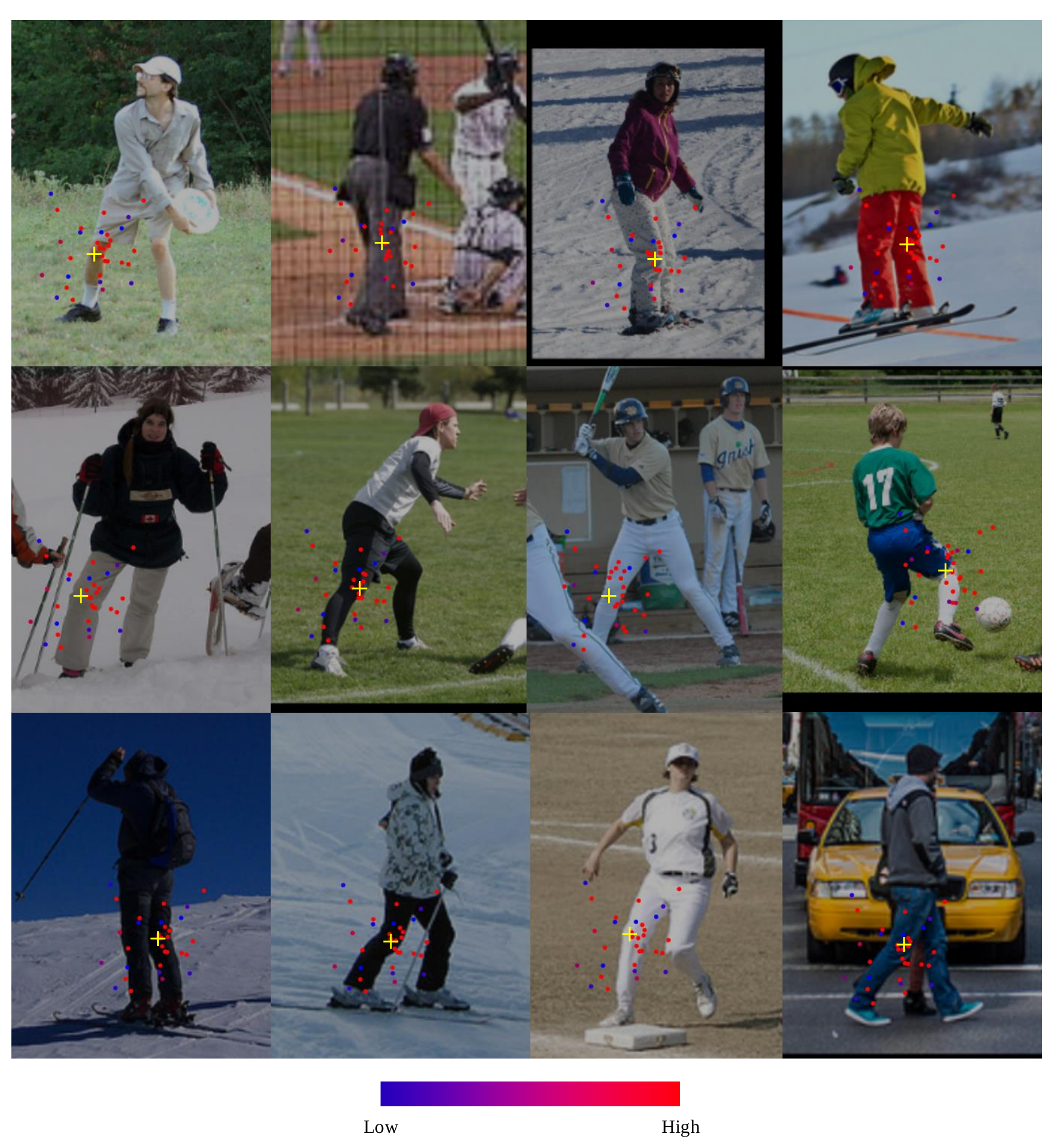}
\caption{Visualization of right knee's pixel-to-query attention in the last decoder layer. For readability, we draw the sampling points and attention weights from $C_3$ feature map in different pictures. Each sampling point is marked as a filled circle whose color indicates its corresponding weight. The ground truth joint is shown as yellow cross marker.}
\label{fig:right knee}
\vspace{-1.5em}
\end{figure*}

We visualize the learned multi-scale deformable attention modules for better understanding. As shown in Figure~\ref{fig:right shoulder} and Figure~\ref{fig:right knee}, the visualization indicates that TFPose looks at context information surround the ground truth joint. More concretely, the sampling points near the ground truth joint have higher attention weight (denoted as red), while the sampling points far from the ground truth joint own lower attention weight (denoted as blue).

\end{document}